\documentclass[11pt,a4paper]{article}
\usepackage[hyperref]{acl2021}
\usepackage{times}
\usepackage{latexsym}

\usepackage{microtype}

\aclfinalcopy 
\def\aclpaperid{3437} 

\usepackage{amsmath}
\usepackage{amsfonts}
\usepackage{amssymb}
\usepackage{mathbbol}
\usepackage[ruled,vlined]{algorithm2e}
\usepackage{newtxmath}
\usepackage{microtype}
\usepackage{booktabs}
\usepackage{graphicx}
\usepackage{xcolor,colortbl}
\usepackage{pifont}
\usepackage{booktabs}
\usepackage{multirow}
\usepackage{soul}

\usepackage{tcolorbox}
\definecolor{DPLightGreen}{rgb}{0.007, 0.410, 0.246}

\usepackage{tcolorbox}
\definecolor{DPLightRed}{rgb}{0.941, 0.585, 0.551}

\newtcbox{\dpboxblue}{on line,
  colframe=DPLightGreen,colback=DPLightGreen!10!white,
  boxrule=0.5pt,arc=2pt,boxsep=0pt,left=2pt,right=2pt,top=2pt,bottom=2pt}
  
\newtcbox{\dpboxred}{on line,
  colframe=DPLightRed,colback=DPLightRed!10!white,
  boxrule=0.5pt,arc=2pt,boxsep=0pt,left=2pt,right=2pt,top=2pt,bottom=2pt}

\newcommand\redit[1]{{\color{black} #1}}
\newcommand\dpedit[1]{{\color{black} #1}}

\definecolor{RBRed}{rgb}{0.98,0.88,0.85}
\definecolor{RBBlue}{rgb}{0.81,0.80,0.94}
\definecolor{RBGreen}{rgb}{0.85,0.91,0.84}
\definecolor{DPGreen}{rgb}{0,0.45,0.24}

\usepackage{todonotes}
\makeatletter
\newcommand*\iftodonotes{\if@todonotes@disabled\expandafter\@secondoftwo\else\expandafter\@firstoftwo\fi} 
\makeatother

\newcommand\norm[1]{\left\lVert#1\right\rVert}

\title{Language Model Augmented Relevance Score}

\author{Ruibo Liu$^{1}$ \hspace{0.5cm} Jason Wei$^{2}$ \hspace{0.5cm} Soroush Vosoughi$^{1}$\\
  $^{1}$Dartmouth College \hspace{0.5cm} $^{2}$Google AI Language\\
   \texttt{\href{mailto:ruibo.liu.gr@dartmouth.edu}{ruibo.liu.gr@dartmouth.edu}} \hspace{0.2cm} \texttt{\href{mailto:jasonwei@google.com}{jasonwei@google.com}} \\
    \texttt{\href{mailto:soroush.vosoughi@dartmouth.edu}{soroush.vosoughi@dartmouth.edu}}  \\}

\date{}

\begin{document}
\maketitle
\begin{abstract}

Although automated metrics are commonly used to evaluate NLG systems, they often correlate poorly with human judgements.
Newer metrics such as BERTScore have addressed many weaknesses in prior metrics such as BLEU and ROUGE, which rely on $n$-gram matching.
These newer methods, however, are still limited in that they do not consider the generation context, so they cannot properly reward generated text that is correct but deviates from the given reference. 

In this paper, we propose Language \underline{M}odel \underline{A}ugmented \underline{R}elevance \underline{S}core (MARS), a new context-aware metric for NLG evaluation.
MARS leverages off-the-shelf language models, guided by reinforcement learning, to create augmented references that consider both the generation context and available human references, which are then used as additional references to score generated text.
Compared with seven existing metrics in three common NLG tasks, MARS not only achieves higher correlation with human reference judgements, but also differentiates well-formed candidates from adversarial samples to a larger degree.

\end{abstract}

\section{Introduction}

Automated metrics such as BLEU~\cite{papineni2002bleu} and ROUGE~\cite{lin2004rouge} are popular methods for evaluating natural language generation (NLG) systems. Compared with human evaluation, they are cheaper and faster, and accordingly, they often serve as essential metrics for benchmarking the performance of NLG models~\cite{novikova-etal-2017-need}. Despite their widespread use, however, these automated metrics often poorly correlate with ratings given by human judges, particularly for datasets in which only a single human reference exists~\cite{gupta2019investigating,novikova-etal-2017-need}. Moreover, these automated metrics only capture similarities between generated sentences and reference candidates, crucially ignoring provided contexts that are relevant for evaluating the answer in contextual NLG tasks, such as story generation, news summarization, and question-answering~\cite{tao2018ruber,nema-khapra-2018-towards}.

\begin{figure}[!t]
\centering
\includegraphics[width=0.49\textwidth]{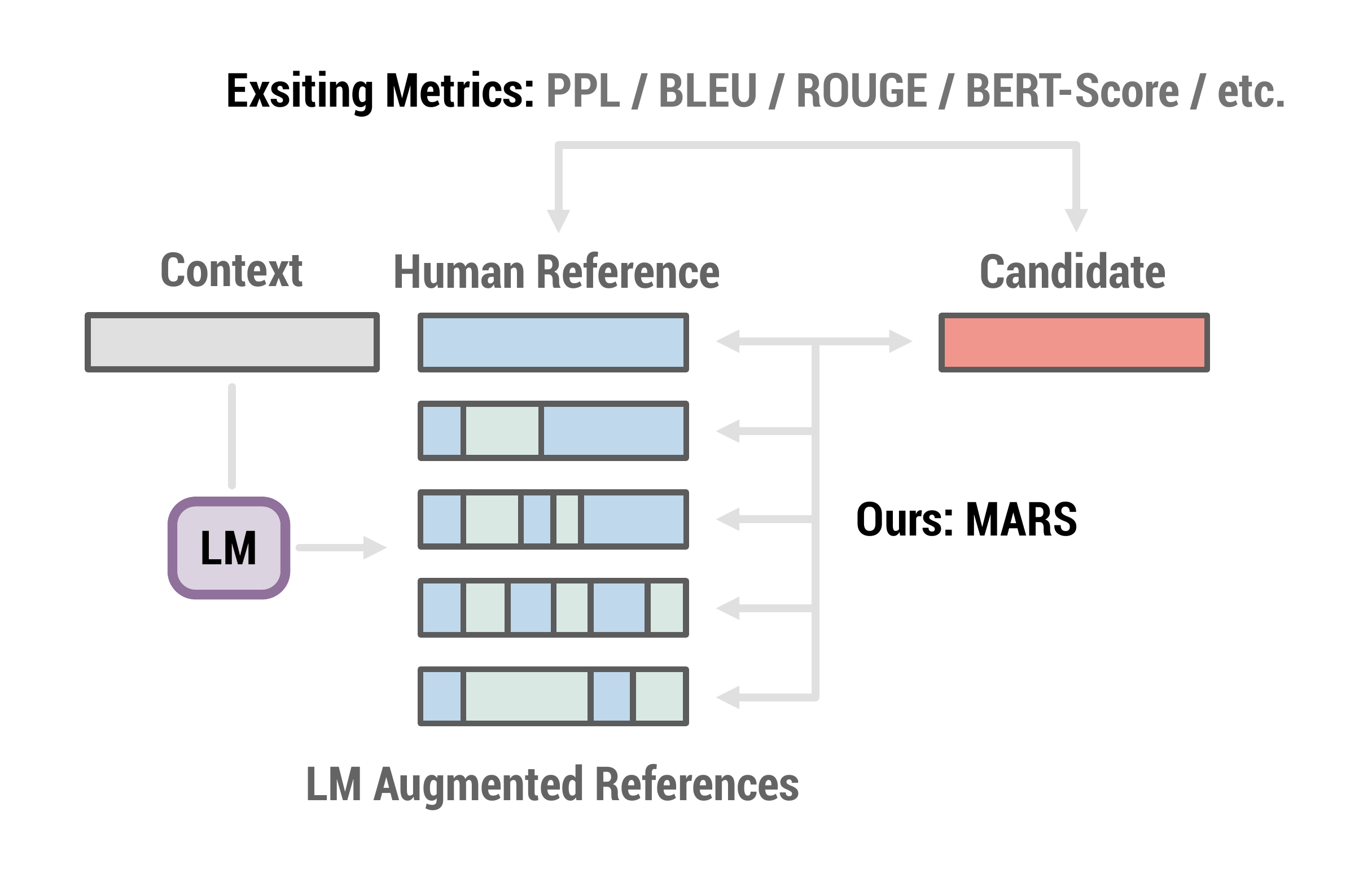}
\vspace{-0.3in}
\caption{
Existing metrics compare the candidate with the human reference but ignore context.
MARS (our method) augments the human reference while considering the context, which allows it to provide evaluation scores that correlate highly with human references.
}
\vspace{-0.15in}
\label{fig:intro}
\end{figure}

\begingroup
\setlength{\tabcolsep}{2.5pt}
\begin{table*}[!ht]
\centering
\resizebox{\textwidth}{!}{%
\begin{tabular}{@{}lccccc@{}}
\toprule
\multicolumn{6}{l}{\begin{tabular}[l]{@{}l@{}}\textbf{Context.} Wendy was driving down the road. She heard her car making a noise. She pulled over to examine the problem.\\There was nothing but oil all on the road from her car.\end{tabular}} \\ \midrule
\textbf{Human Reference.}  She called for help and waited to get her car fixed.           & \textbf{PPL}    & \textbf{BLEU-1} & \textbf{ROUGE-L} & \textbf{BERTScore} & \textbf{MARS}  \\ \midrule
\textbf{Candidate.} Her fears were confirmed when her engine was smoking.      & \dpboxred{75.58}  & \dpboxred{0.223}   & \dpboxred{0.182}    & \dpboxred{0.338}      & 0.574 \\
\textbf{Reorder.} her confirmed engine fears Her when was were smoking.         & 405.60 & \dpboxred{0.223}   & \dpboxred{0.182}    & 0.265      & 0.352 \\

\textbf{Retrieve.} She heard her car making a noise.                   & \dpboxred{63.93}  & 0.337   & 0.400    & \dpboxred{0.406}      & 0.448 \\ \bottomrule
\end{tabular}%
}
\caption{
In this story generation example, MARS is the only metric that gives the well-formed candidate a higher score than two adversarial examples. 
The human rating of the candidate averaged over 20 judgements is 5.05 out of 6.00.
Two adversarial examples are generated by \textbf{Reorder}ing the tokens of the candidate \redit{(as weak NLG systems whose generation is not readable) and \textbf{Retrieve}ing a sentence from the context (as systems with no generation ability).}
We \dpboxred{boxed} the cases where the adversarial example does not score lower than the well-formed candidate.
}
\vspace{-0.15in}
\label{tab:demo}
\end{table*}
\endgroup


Table~\ref{tab:demo} shows a story generation\footnote{\redit{The ROC story generation task asks systems to generate a legitimate ending for a four-sentence story.}} example that exemplifies some weaknesses of several common metrics.
Perplexity (PPL)~\cite{brown1992estimate} successfully detects ungrammatical sentences, but it fails to distinguish legitimate novel continuations and copy-and-pasted ones. 
Relying on surface-level $n$-gram matching, BLEU-1 and ROUGE-L\footnote{L stands for longest common sequence matching.} \redit{cannot detect reordering effectively, and wrongly score the well-formed candidate lower than its retrieval-based adversarial example.}
BERTScore~\cite{zhang2019bertscore} leverages contextual embeddings from BERT~\cite{devlin2019bert}, thus mitigating the above challenges, but still does not fairly evaluate candidates that correctly align with the context but happen to differ from the provided reference example. In our example, the candidate \textit{``... her engine was smoking''} is reasonable but deviates from the human reference, and so BERTScore rates it relatively low (0.338 out of 1.0), thus correlating poorly with human rating, which was high (5.05 out of 6.00).

To address the above issues, prior studies have proposed a number of promising remedies. One line of work has proposed to combine human ratings with automated metrics~\cite[][\textit{inter alia}]{durmus2020feqa,chaganty-etal-2018-price}. 
For instance, in HUSE score, \citet{hashimoto-etal-2019-unifying} leverages the differences between perplexity and human judgements to consider both quality and diversity of generated text.
Another line has proposed training separate neural models to aid automated metrics~\cite[][\textit{inter alia}]{mehri-eskenazi-2020-usr,yuma-etal-2020-ubleu}. For instance, BLEURT~\cite{sellam-etal-2020-bleurt} fine-tunes BERT~\cite{devlin2019bert} on synthetic reference-candidate pairs for machine translation. These methods, however, are often limited in practical use, because the high-cost human ratings are not always available for every dataset, and the data- or system-specific training is not easily extended to other domains~\cite{zhang2019bertscore}, \redit{and can even bias the evaluation~\cite{freitag-etal-2020-bleu}.}

In this paper, we present MARS (Language \underline{M}odel \underline{A}ugmented \underline{R}elevance \underline{S}core), a new NLG evaluation metric that requires neither supervision from human ratings nor additional training on specific domains. As shown in Figure~\ref{fig:intro}, instead of comparing candidates only with human written references, as many prior metrics do, MARS 
 
uses a mixture of both human and augmented references. Specifically, MARS masks tokens in the reference to create templates, and then uses the context and templates to generate augmented references by infilling the masked parts with an LM guided by reinforcement learning. 
The augmented references thus incorporate information from both the context and the human reference, and are enriched with lexical and syntactic diversity, facilitating fairer evaluation of candidates.
Finally, we compute the score as a weighted average of the similarity between the candidate and the set of augmented references in the contextual embedding space.

The advantages of MARS are three-fold. \textit{First}, MARS correlates highly with human judgements. We apply MARS to three diverse NLG tasks, and demonstrate that, compared with seven popular NLG metrics, MARS better correlates with human judgements and is robust against adversarial attacks. 
\textit{Second}, MARS is context-aware. Unlike existing metrics that only consider the given human reference, we use a constrained NLG approach to incorporate the generation context into augmented references, thus alleviating bias against diverse candidates. 
\textit{Third}, MARS is easy to deploy and extend. Built on off-the-shelf LMs, MARS requires neither human supervision nor additional training for specific domains, and can therefore serve as a general-purpose metric for a broad range of NLG applications, as we will demonstrate for three common NLG tasks: story generation, news summarization, and question-answering.

\section{Approach}

MARS comprises three steps. First, we mask out non-important tokens from the human reference to produce templates for augmentation (\S\ref{subsec:token_masking}).
Second, we guide off-the-shelf LMs to generate reference augmentation on these templates via a reinforced self-planning algorithm (\S\ref{subsec:self_planning}).
Finally, we compute a weighted average score that reflects the overall similarity between the candidate and the set of augmented references (\S\ref{subsec:contextual_similarity}).

\subsection{Human Reference Token Masking}
\label{subsec:token_masking}

The first step in MARS is to take in the given human reference and generate \textit{templates}---masked versions of the human reference---which can then be used to generate augmented references.
Our masking procedure can be viewed as a reversed process of prior insertion- and template-based generation approaches~\cite{zhang-etal-2020-pointer,miao2019cgmh}; whereas these generation approaches start with templates of important tokens and then fill in the details to generate complete sentences, our masking procedure starts with the complete sentence (i.e., the human reference) and then masks out unimportant tokens to generate templates. 

To better explain our masking procedure, we introduce two concepts, mask priority and mask cost: 

\paragraph{Mask Priority.} We compute a mask priority $v_i$ for each token $x_i$, which captures the priority of masking $x_i$, where non-important words should receive higher priority. 
We compute $v_i$ as a function of two things: the inverse document frequency (IDF) of $x_i$, and the part-of-speech (POS) of $x_i$: 
\begin{equation}
    v_i = \frac{\alpha(\textrm{POS}\left[x_i\right])}{\textrm{IDF}(x_i, X)} \ ,
\end{equation}
where $\alpha$ is a function that assigns a weight to each POS tag.\footnote{$\alpha$ varies for each task. Empirically, we find that it works well to assign adjectives, adverbs, and nouns higher weights than other parts-of-speech. For our setting, we assign weights of 4, 3, 2 to the above three types.}
Common tokens across the corpus $X$ (e.g., stop words, with low IDF) will receive high mask priority. Tokens responsible for description details will also be assigned high mask priority based on their part-of-speech (e.g., adjectives are mainly used for details and so they are given higher priority of being masked).

\paragraph{Mask Cost.} For each token $x_i$, we also compute a mask cost $w_i$. Tokens that appear in both context and human reference should have high masking cost as they are deemed context-carrying. We use the longest common sequence (LCS) matching between the context and the human reference to identify these context-carrying tokens. In our experiments, we set the $w_i$ of these tokens to 10 and the default $w_i$ of all other tokens to 1. We use $\lambda$ to denote the ratio of tokens to be masked in a sentence of $N$ tokens, and define $W_{\textrm{max}} = \lambda \cdot N$ as the maximum cost allowed.

\paragraph{DP-based Token Masking.}
Now that for each token we have a mask priority and a mask cost, we aim to choose a set of tokens to mask with the highest possible sum of priorities for which the sum of mask costs is not greater than
$W_{\textrm{max}}$. Given a function $\phi(x_i) = \{1, 0\}$ where 1 means token $x_i$ is masked and 0 means it remains, the objective of token masking can be expressed as follows:

\begin{equation}
\label{eqa:token_masking}
\begin{aligned}
    \textrm{max} &\sum_{i=1}^{N} v_i \cdot \phi(x_i)\ , 
    \\ \textrm{s.t.}\ &\sum_{i=1}^{N} w_i \cdot \phi(x_i) \leq W_{\textrm{max}}\ .
\end{aligned}
\end{equation}

Such a goal is actually a NP-complete combinatorial optimization problem, called the Knapsack problem~\cite{pisinger1995algorithms}, which we solve using dynamic-programming (DP). 
In general, the masking strategy aggressively harvests tokens of high mask priority while keeping the cost of masked tokens from exceeding the mask cost limitation $W_{\textrm{max}}$. \dpedit{The detailed DP algorithm for solving this problem is shown in Appendix A.}

\subsection{Self-planning \textit{Cloze} Augmentation} 
\label{subsec:self_planning}

After creating the templates described in $\S$\ref{subsec:token_masking}, we produce augmented reference examples based on both the templates as well as the generation context. This procedure can be seen as a mixture of hard- and soft-constrained NLG, where the template tokens pre-exist with some blanks, and the system, conditioned on the context, aims to fill in the blanks. We henceforth refer this process of creating augmented references as \textit{cloze}\footnote{A \textit{cloze} test~\cite{taylor1953cloze} is a language test where a portion of language is removed and the participant is asked to replace the missing language item.} augmentation.

\paragraph{Background.} Masked Language Models (MLM) such as RoBERTa~\cite{liu2019roberta} and BERT~\cite{devlin2019bert} are trained to predict masked tokens within sentences, and thus are able to do \textit{cloze} augmentation off-the-shelf. However, without architecture-level modification, MLMs are only able to infill a \textit{pre-determined} number of missing tokens~\cite{zhu2019text}. This is especially problematic since---if they are directly used to augment references---all the augmented references will have the same number of tokens as that of the original human reference. 
We believe this unnecessarily constrains augmentation diversity, and thus consider it as a Naive method in our evaluations ($\S$\ref{sec:evaluation}).

\begin{figure}[!ht]
\centering
\includegraphics[width=0.49\textwidth]{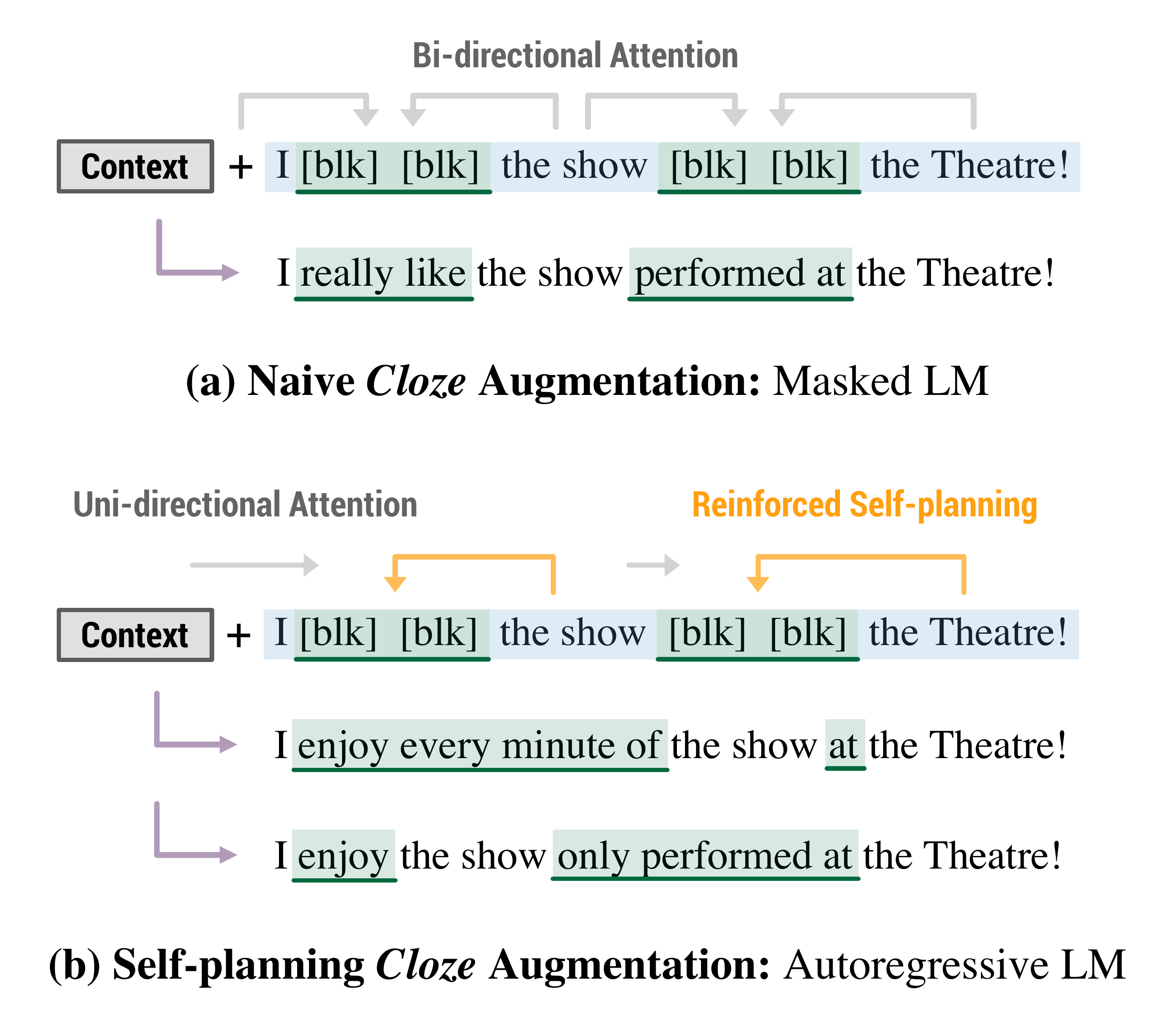}
\vspace{-0.25in}
\caption{Compared with the Naive method, our reinforced self-planning approach infills blanks with ([blk]) varying-length tokens while considering both past and future tokens, which promote diversity and coherence respectively. 
The context is concatenated to the beginning of the reference template. }
\label{fig:self_planning}
\vspace{-0.15in}
\end{figure}

Autoregressive Language Models (ALM) such as GPT-2~\cite{radford2019language}, on the other hand, are trained to predict current step token given past tokens. They can generate sequences of \textit{varying} lengths, but they cannot infill missing tokens within sentences effectively since they do not consider future context. To enable ALMs to infill blanks of unspecified length, prior work has proposed either retraining a new LM from scratch~\cite{shen-etal-2020-blank} or fine-tuning on specially prepared data~\cite{donahue-etal-2020-enabling}, which are costly and not easy to extend to new NLG tasks. As shown in Figure~\ref{fig:self_planning}, we take a reinforcement learning (RL) approach that uses future words after the blank to guide current step infilling generation. \redit{Since such RL guidance only relies on the tokens within its own to-be-infilled template, we call it reinforced \textit{self-planning}. Our method combines the advantages of both MLMs and ALMs, requiring neither re-training nor collecting new data, and thus is easier to extend to other off-the-shelf LMs.}

\paragraph{Reinforced Self-planning.} At each decoding step during generation, a vanilla ALM will pick the token $x_t$ that has the highest probability by applying an argmax over the softmax output of hidden states. We add a self-planning stage between the argmax and softmax function. Following the RL framework, we define the \textit{state} at step $t$ as the generated sequences before $t$ (i.e., $s_t = x_{<t}$), and the \textit{action} at step $t$ as the $t$-th output token (i.e., $a_t = x_{t}$). We take the softmax output of the last hidden states (with parameter $\theta$) as the policy $\pi_{\theta}$, since it is the probability of picking token $x_t$ (action $a_t$) given the state $s_t = x_{<t}$. Similarly, we denote the policy after reinforced self-planning as $\pi_{\theta_d}$.

Typically, the RL objective is to maximize the expectation of total reward $J$, summed over $T$ steps on the trajectory $\tau$ induced by $\pi_{\theta}$:
\begin{equation}
\label{eqa:rewards}
    J(\theta) = \mathbb{E}_{\tau \sim \pi_{\theta}}\left[\sum_{t=0}^{T} \gamma^t r_t \right],
\end{equation}
where $\gamma \in (0, 1]$ is the discounting factor, and $r$ is the single-step reward. In text generation, however, such a reward definition requires sampling over the future generated sequence to estimate current step reward~\cite{gong-etal-2019-reinforcement}, which may cause the policy to end in zero reward region because of high variance of the gradient~\cite{pang2020text}. Since we guide the generation in every step of decoding, we derive the $t$-th step policy gradient $\triangledown_{\theta} J_t(\theta)$ as:
\begin{equation}
\label{eqa:gradient}
    \mathbb{E}^t_{\tau \sim \pi_{\theta}}\left[\epsilon_t \triangledown_{\theta} \log \pi_{\theta}(a_t|s_t) \cdot r(x_t^d)\right]\ ,
\end{equation}
with importance sampling weight $\epsilon_t$ to stabilize the optimization~\cite{munos2016safe}, which is:
\begin{equation*}
    \epsilon_t = \frac{\pi_{\theta_{d}}(a_t|s_t)}{\pi_{\theta}(a_t|s_t)}\ .
\end{equation*}

If we denote a certain token in future context as $w \in \{w_{\textrm{future}}\}$, single-step self-planning reward $r(x_t^d)$ can be approximated by the cosine similarity between $t$-th step hidden state and the embedded vector of $w$ by the LM embedding layers, which is
\begin{equation}
\label{eqa:self_planning_reward}
    r(x_t^d) = \sum_{w \in w_{\textrm{future}}} \log (\textrm{softmax}(h_{<t}^{\theta_d}) \cdot \textrm{emb}(w))\ .
\end{equation}

Given all above definitions, at $t$-th step, we update $\pi_{\theta}$ towards the self-planned $\pi_{\theta_d}$ as:
\begin{equation}
\label{eqa:policy_update}
    \theta_d \leftarrow \theta + \eta \sum_{i=1}^k \frac{\triangledown_{\theta} J_t(\theta_d / \xi) }{\norm{\triangledown_{\theta} J_t(\theta_d / \xi)}}\ ,
\end{equation}
where $\eta$ is the learning rate and $\xi$ is the temperature parameter to control the stochastic sampling during token decoding~\cite{keskar2019ctrl}. After $k$ iterations of reinforced self-planning, the updated policy $\pi_{\theta_d}$ should produce tokens approaching the future context in embedding space, since future context contributes to the calculation of reward $r$ (Eq.~\ref{eqa:self_planning_reward}).\footnote{In our setting, $\eta$, $\xi$ and $k$ are 0.02, 1.3, and 3 respectively.} More details about how we handle edge cases during reinforced self-planning are presented in Appendix B.

\subsection{Computing Contextual Similarity}
\label{subsec:contextual_similarity}

After generating augmented reference sentences, the final MARS score is computed as a weighted average of the similarity between the candidate and each reference in the augmentation set (including the original human reference).
One way to obtain similarity scores is using BERTScore~\cite{zhang2019bertscore}, but BERTScore requires training on external resources to make its outputs more readable.
Therefore, in order to keep all the resources used by MARS off-the-shelf, we utilize Sentence-BERT~\cite{reimers-2019-sentence-bert}, which uses the mean of all token embeddings in a sentence as the overall sentence-level encoding. 
As the sentence encoder, we use RoBERTa-large~\cite{liu2019roberta}, a common choice in the literature~\cite{zhang2019bertscore,reimers-2020-multilingual-sentence-bert}. 
As shown in Eq.~\ref{eqa:mars_score}, we then compute MARS score as the average of the cosine similarities weighted using a geometric progression with a common ratio $q \in (0, 1]$ and a scale factor (start value) $a \neq 0$:
\begin{equation}
\label{eqa:mars_score}
\begin{aligned}
    \textrm{MARS} = &\sum_{i=1}^{\#\lambda} aq^{i-1} \frac{\textrm{cand}^{\textrm{T}}\cdot \textrm{ref}_{i-1}}{\norm{\textrm{cand}}^{\textrm{T}} \norm{\textrm{ref}_{i-1}}}
    \\ \textrm{s.t.}\ &\sum_{i=1}^{\#\lambda} aq^{i-1} = 1 \ ,
\end{aligned}
\end{equation}
where the candidate encoding is cand, the reference encodings are $\textrm{ref}_{i}$ ($i$ is the index of the augmented reference under a certain $\lambda$, and $\textrm{ref}_0$ marks the zero-mask human reference), and $\#\lambda$ is the number of masking ratios we use in $\S$\ref{subsec:token_masking}.
Different $q$ values, as defined by the geometric progression, determine how much weight each reference contributes. By default, Eq.~\ref{eqa:mars_score} assigns the largest weight to the human reference since it is the gold standard. 

\section{Tasks \& Datasets}
\label{sec:tasks_datasets}

We evaluated MARS and compared it with several popular NLG metrics on the following three tasks:

\vspace{0.1in}

\noindent \textbf{Story Generation.} We use the ROC stories dataset\footnote{https://cs.rochester.edu/nlp/rocstories/} for story generation, which requires candidate NLG systems to generate coherent endings to four-sentence stories~\cite{mostafazadeh-etal-2016-corpus}. The dataset consists of 96,198 examples of partially written stories; we take the human-rated subset ($N$=300) released by HUSE~\cite{hashimoto-etal-2019-unifying}, which contains continuances by (1) an industry-level system based on Apache \textbf{Solr}\footnote{https://lucene.apache.org/solr}, and (2) an \textbf{Open-NMT} model with global attention~\cite{mccann2017learned}.

\vspace{0.05in}

\noindent \textbf{News Summarization.} For the news summarization task, we use the Newsroom summary dataset.\footnote{http://lil.nlp.cornell.edu/newsroom/} This dataset contains 1.3 million articles from 38 major publications~\cite{grusky-etal-2018-newsroom} and we use the subset with human ratings ($N$=540) released by the authors.\footnote{The subset includes human ratings on four perspectives: \textit{coherence}, \textit{fluency}, \textit{informative} and \textit{relevance}. We compute the average of the four scores as an overall human rating.} This dataset contains outputs from summarization models: (1) \textbf{TextRank}: a sentence-level summarization system inspired by Google PageRank~\cite{page1999pagerank}, (2) a \textbf{Seq2Seq} model with attention~\cite{rush-etal-2015-neural}, and (3) \textbf{Pointer-N}: a pointer-based neural model~\cite{see2017get} trained on Newsroom dataset.

\vspace{0.05in}

\begingroup
\setlength{\tabcolsep}{3.5pt}
\begin{table}[!t]
\centering
\resizebox{0.47\textwidth}{!}{%
\begin{tabular}{@{}lccccc@{}}
\toprule
 &
  \begin{tabular}[c]{@{}c@{}}\textit{Avg.}\\ $\lvert$Cntx.$\lvert$ \end{tabular} &
  \begin{tabular}[c]{@{}c@{}}\textit{Avg.} \\ $\lvert$H Ref.$\lvert$\end{tabular} &
  $\Omega$ &
  \begin{tabular}[c]{@{}c@{}}\# data\\ (\# HR / data)\end{tabular} &
  $\alpha$ \\ \midrule
\textbf{ROC}      & 34.38  & 8.37  & 4.1  & 300 (20) & 0.64  \\
\textbf{Newsroom} & 772.21 & 34.70 & 22.3 & 540 (3)  & 0.71  \\
\textbf{MOCHA}    & 161.92 & 4.69  & 34.5 & 450 (5)  & 0.82 \\ \bottomrule
\end{tabular}%
}
\caption{Statistics of the three datasets with human ratings used in this work. Avg. $\lvert$Cntx.$\lvert$ and $\lvert$H Ref.$\lvert$: the averaged number of tokens in contexts and human references. $\Omega$: the ratio of the previous two terms (lower $\Omega$ can indicate a more open-ended task). \# HR: the number of Human Ratings. $\alpha$: Krippendorff's alpha coefficient to measure inter-annotator agreement.}
\label{tab:datasets}
\vspace{-0.2in}
\end{table}
\endgroup
\begin{table*}[!ht]
\centering
\resizebox{0.97\textwidth}{!}{%
\begin{tabular}{@{}lcccccccc@{}}
\toprule
\multicolumn{1}{c}{} &
  \multicolumn{2}{c}{\cellcolor{RBRed}\begin{tabular}[c]{@{}c@{}}\textbf{ROC Story Generation}\\ $\Omega$ = 4.1\end{tabular}} &
  \multicolumn{3}{c}{\cellcolor{RBGreen}\begin{tabular}[c]{@{}c@{}}\textbf{Newsroom Summarization}\\ $\Omega$ = 22.7\end{tabular}} &
  \multicolumn{3}{c}{\cellcolor{RBBlue}\begin{tabular}[c]{@{}c@{}}\textbf{MOCHA Question Answering}\\ $\Omega$ = 34.5\end{tabular}} \\ \midrule
\textbf{Existing Metrics} & Solr                      & Open-NMT & TextRank & Seq2Seq & Pointer-N & GPT-2  & Back-Tran & MHPG   \\ \cmidrule(l){2-9} 
BLEU-1                    & 0.198                     & 0.104    & 0.224    & 0.268   & 0.115     & 0.328  & 0.061     & 0.318  \\
METEOR                    & 0.180                     & 0.116    & 0.288    & 0.235   & 0.256     & 0.466  & 0.179     & 0.409  \\
ROUGE-L                   & 0.118                     & 0.195    & 0.041    & -0.133  & 0.065     & 0.468  & 0.056     & 0.247  \\
Sent.~Mover Sim.           & 0.020                     & 0.015    & 0.112    & 0.099   & 0.177     & 0.510  & 0.166     & 0.610  \\
MoverScore               & 0.181                     & 0.391    & 0.075    & \textbf{0.337}   & 0.212     & \textbf{0.535}  & 0.190     & 0.592  \\
BERTScore                 & 0.245                     & 0.386    & 0.154    & 0.302   & 0.181     & 0.444  & 0.274     & 0.458  \\
Perplexity               & -0.104                    & -0.073   & -0.385   & 0.011   & -0.035    & 0.014  & -0.051    & -0.128 \\ \cmidrule(r){1-1}
\textbf{MARS} (default)             & \textbf{0.476} &                    \textbf{0.397}    & \textbf{0.372}    & 0.336   & \textbf{0.329}     & 0.526  & \textbf{0.644}     & \textbf{0.741}  \\
- \ w/o. self-plan.            & 0.313                     & 0.212    & 0.290    & 0.245   & 0.314     & 0.477  & 0.631     & 0.709  \\
- \ w/o. context$^+$              & 0.360                     & 0.334    & 0.107    & 0.160   & -0.009    & 0.134  & 0.222     & 0.303  \\
- \ w/o. \textit{both}                 & \multicolumn{1}{c}{0.276} & 0.183    & -0.163   & 0.149   & -0.057    & -0.092 & 0.121     & 0.299  \\ \cmidrule(r){1-1}
Naive (MLM)                    & 0.449                     & 0.197    & 0.201    & 0.324   & 0.114     & 0.443  & 0.307     & 0.540  \\ \bottomrule
\end{tabular}%
}
\caption{
Pearson's $r$ correlations with human judgements for MARS and seven existing metrics across system outputs for three generation tasks.
BLEU-1~\cite{papineni2002bleu}, METEOR~\cite{lavie-agarwal-2007-meteor}, and ROUGE-L~\cite{lin-och-2004-automatic} use $n$-gram matching. 
Sentence Mover's Similarity~\cite{clark-etal-2019-sentence} and MoverScore~\cite{zhao-etal-2019-moverscore} measure similarity using earth mover's distance.
BERTScore~\cite{zhang2019bertscore} leverages contextual embeddings from pre-trained LMs. 
As an ablation, we remove self-planning guidance, context, and both. Naive uses RoBERTa-large for reference augmentation (see $\S$\ref{subsec:self_planning}). $\Omega$ is defined as in Table~\ref{tab:datasets}.
}
\vspace{-0.05in}
\label{tab:correlation}
\end{table*}

\noindent\textbf{Question Answering.} For question answering, we use the MOCHA dataset,\footnote{https://allennlp.org/mocha} which includes human ratings on outputs of five models trained on six QA datasets~\cite{chen2020mocha}. We consider a distributionally-balanced subset ($N$=450) of these outputs from three systems: (1) fine-tuned \textbf{GPT-2}~\cite{radford2019language}, (2) a \textbf{Back-Translation} model~\cite{sennrich2016improving}, and (3) a \textbf{MHPG} model~\cite{bauer2018commonsense} trained on NarrativeQA~\cite{kovcisky2018narrativeqa} and MCScript~\cite{ostermann2018mcscript} datasets.

\vspace{0.05in}

The detailed statistics of these three datasets we used for this work are shown in Table~\ref{tab:datasets}. For pre-processing, we removed hashtags and urls in the text, but leave punctuation and stop words, which can affect LCS matching when computing mask costs. 
For all tasks, we use GPT-2 (large, with 774M parameters) as the language model for MARS, and RoBERTa-large for the Naive method.
For the newsroom dataset, some news articles were longer than the max sequence length of 1024 BPE, and so we cut off the tail end of these examples.
With a single RTX-2080 GPU, \textit{cloze} augmentation with $\lambda$ = \{0 (human \textit{ref.}), 20\%, 40\%, 60\%, 80\%\} takes 0.8 seconds on average per reference, amounting to a total augmentation time of 17, 45, and 32 minutes for the ROC, Newsroom and MOCHA tasks respectively. We show how we pick the masking ratios for different tasks in $\S$\ref{subsec:best_masking}.

\section{Evaluation}
\label{sec:evaluation}

\subsection{MARS Better Correlates With Humans}
\label{subsec:correlation}
As automated metrics are only helpful if they correlate sufficiently with human judgements, in this section we examine how MARS correlates with human judgements compared with prior metrics.

\vspace{-1mm}
\paragraph{System-level Correlation.}

\begingroup
\setlength{\tabcolsep}{3pt}
\begin{table*}[!ht]
\centering
\resizebox{\textwidth}{!}{%
\begin{tabular}{@{}lccccccccc@{}}
\toprule
\multicolumn{1}{c}{} &
  \multicolumn{3}{c}{\cellcolor{RBRed}\textbf{ROC Story Generation}} &
  \multicolumn{3}{c}{\cellcolor{RBGreen}\textbf{Newsroom Summarization}} &
  \multicolumn{3}{c}{\cellcolor{RBBlue}\textbf{MOCHA Question Answering}} \\ \cmidrule(l){2-10} 
\textbf{Existing Metrics} &
  Reorder ($\Delta$)  &
  Retrieve ($\Delta$)  &
  \textit{ref.} &
  Reorder ($\Delta$)  &
  Retrieve ($\Delta$) &
  \textit{ref.} &
  Reorder ($\Delta$) &
  Retrieve ($\Delta$) &
  \textit{ref.} \\ \midrule

BLEU-1          & (\textcolor{red}{=}) 0     & \textcolor{DPGreen}{$\blacktriangledown$} 0.015 & 0.137 & (\textcolor{red}{=}) 0      & \textcolor{DPGreen}{$\blacktriangledown$} 0.144 & 0.176 & (\textcolor{red}{=}) 0     & \textcolor{DPGreen}{$\blacktriangledown$} 0.424 & 0.344 \\

METEOR          & \textcolor{DPGreen}{$\blacktriangledown$} 0.041     & \textcolor{DPGreen}{$\blacktriangledown$} 0.031 & 0.094 & \textcolor{DPGreen}{$\blacktriangledown$} 0.132      & \textcolor{DPGreen}{$\blacktriangledown$} 0.142 & 0.244 & \textcolor{DPGreen}{$\blacktriangledown$} 0.012     & \textcolor{DPGreen}{$\blacktriangledown$} 0.379 & 0.412 \\

ROUGE-L         & \textcolor{DPGreen}{$\blacktriangledown$} 0.131 & \textcolor{DPGreen}{$\blacktriangledown$} 0.123 & 0.194 & \textcolor{red}{$\blacktriangle$} 0.011 & \textcolor{DPGreen}{$\blacktriangledown$} 0.035 & 0.036 & \textcolor{DPGreen}{$\blacktriangledown$} 0.032 & \textcolor{DPGreen}{$\blacktriangledown$} 0.363 & 0.336 \\


Sent. Mover Sim.         & \textcolor{DPGreen}{$\blacktriangledown$} 0.024 & \textcolor{DPGreen}{$\blacktriangledown$} 0.062 & 0.019 & \textcolor{DPGreen}{$\blacktriangledown$} 0.153 & \textcolor{DPGreen}{$\blacktriangledown$} 0.161 & 0.136 & \textcolor{DPGreen}{$\blacktriangledown$} \underline{0.232} & \textcolor{DPGreen}{$\blacktriangledown$} 0.161 & 0.515 \\

MoverScore         & \textcolor{DPGreen}{$\blacktriangledown$} 0.131 & \textcolor{DPGreen}{$\blacktriangledown$} 0.123 & 0.276 & \textcolor{red}{$\blacktriangle$} 0.011 & \textcolor{DPGreen}{$\blacktriangledown$} 0.135 & 0.236 & \textcolor{red}{$\blacktriangle$} 0.027 & \textcolor{DPGreen}{$\blacktriangledown$} 0.495 & 0.500 \\

BERTScore      & \textcolor{DPGreen}{$\blacktriangledown$} 0.109 & \textcolor{DPGreen}{$\blacktriangledown$} 0.127 & 0.337 & \textcolor{DPGreen}{$\blacktriangledown$} 0.112  & \textcolor{DPGreen}{$\blacktriangledown$} 0.026 & 0.344 & \textcolor{DPGreen}{$\blacktriangledown$} 0.101 & \textcolor{DPGreen}{$\blacktriangledown$} 0.461 & 0.462 \\
Perplexity      & \textcolor{DPGreen}{$\blacktriangledown$} 0.113 & \textcolor{red}{$\blacktriangle$} 0.170 & -0.089 & \textcolor{DPGreen}{$\blacktriangledown$} \underline{0.298}  & \textcolor{red}{$\blacktriangle$} 0.008 & 0.234 & \textcolor{DPGreen}{$\blacktriangledown$} 0.035 & \textcolor{red}{$\blacktriangle$} 0.026 & -0.032 \\

\cmidrule(r){1-1}
\textbf{MARS} &
  \multicolumn{1}{l}{} &
  \multicolumn{1}{l}{} &
  \multicolumn{1}{l}{} &
  \multicolumn{1}{l}{} &
  \multicolumn{1}{l}{} &
  \multicolumn{1}{l}{} &
  \multicolumn{1}{l}{} &
  \multicolumn{1}{l}{} &
  \multicolumn{1}{l}{} \\
w/. RoBERTa Emb. & \textcolor{DPGreen}{$\blacktriangledown$} 0.125 & \textcolor{DPGreen}{$\blacktriangledown$} \underline{0.191} & \textbf{0.459} & \textcolor{DPGreen}{$\blacktriangledown$} 0.117  & \textcolor{DPGreen}{$\blacktriangledown$} \underline{0.198} & \textbf{0.423} & \textcolor{DPGreen}{$\blacktriangledown$} 0.092 & \textcolor{DPGreen}{$\blacktriangledown$} \underline{0.504} & \textbf{0.667} \\
w/. GloVe Emb.   & \textcolor{DPGreen}{$\blacktriangledown$} 0.087 & \textcolor{DPGreen}{$\blacktriangledown$} 0.177 & 0.363 & \textcolor{DPGreen}{$\blacktriangledown$} 0.052  & \textcolor{DPGreen}{$\blacktriangledown$} 0.149 & 0.409 & \textcolor{DPGreen}{$\blacktriangledown$} 0.085 & \textcolor{DPGreen}{$\blacktriangledown$} 0.426 & 0.602 \\
\cmidrule(r){1-1}
Naive (MLM)           & \textcolor{DPGreen}{$\blacktriangledown$} \underline{0.149} & \textcolor{DPGreen}{$\blacktriangledown$} 0.156 & 0.350 & \textcolor{DPGreen}{$\blacktriangledown$} 0.112  & \textcolor{DPGreen}{$\blacktriangledown$} 0.190 & 0.314 & \textcolor{DPGreen}{$\blacktriangledown$} 0.098 & \textcolor{DPGreen}{$\blacktriangledown$} 0.247 & 0.639 \\ \bottomrule
\end{tabular}%
}
\caption{We test robustness of MARS and seven other automated metrics under attacks from adversarial samples generated by following two attack strategies: (1) Reorder: randomly reorders 50\% of tokens in the candidates; (2) Retrieve: randomly retrieves a sentence from the context as a candidate. \textit{ref.}: correlation of original candidates with human judgements. If a metric scores adversarial samples equal to (\textcolor{red}{=}) or higher (\textcolor{red}{$\blacktriangle$}) than \textit{ref.}, we consider such metrics not robust under attacks. Robust systems should assign decreased scores (\textcolor{DPGreen}{$\blacktriangledown$}) compared to \textit{ref.}}
\vspace{-0.05in}
\label{tab:robustness}
\end{table*}
\endgroup
Table~\ref{tab:correlation} shows the correlations between human judgements and automated metrics for MARS and seven other unsupervised metrics, across all NLG systems studied in our three tasks. Compared with the other metrics, MARS achieves the highest correlation with human judgements for five of the seven systems (and comparable with the top in the other two systems), making considerable improvements over the next-best metric for many of the NLG systems (e.g., 0.370~$\uparrow$ for Back-Translation, and 0.231~$\uparrow$ for Solr). We also notice that MARS has greater improvements on more open-ended tasks (e.g., story generation, which has low $\Omega$), which corroborates MARS's original objective of judging diverse candidates more fairly. 
As for the baselines, $n$-gram matching metrics such as BLEU correlate poorly with human ratings on such open-ended tasks; BERTScore performs better on short candidates and high-$\Omega$ tasks (e.g., QA); and 
perplexity, as expected, correlates weakly with human ratings. The Naive method, which uses multiple augmented references of the same length, improves over BERTScore, which only uses the original reference. 
\paragraph{Ablation Study.} As shown in the lower rows of Table \ref{tab:correlation}, we see that the performance of MARS drops substantially when the crucial components are removed. 
\redit{Specifically, removing self-planning hurts performance more for tasks with longer references (e.g., story generation)
since self-planning is more helpful when there are more blanks to in-fill, and removing context hurts performance more in tasks that are less open-ended (high $\Omega$, such as QA) because there is no adequate input for a reasonable augmentation. We take these ablation study results as evidence that the techniques we propose in MARS are crucial for improving correlation with human judgements.}

\paragraph{Task-level Correlation Visualization.}

\begin{figure}[!t]
  \centering
   \includegraphics[width=0.49\textwidth]{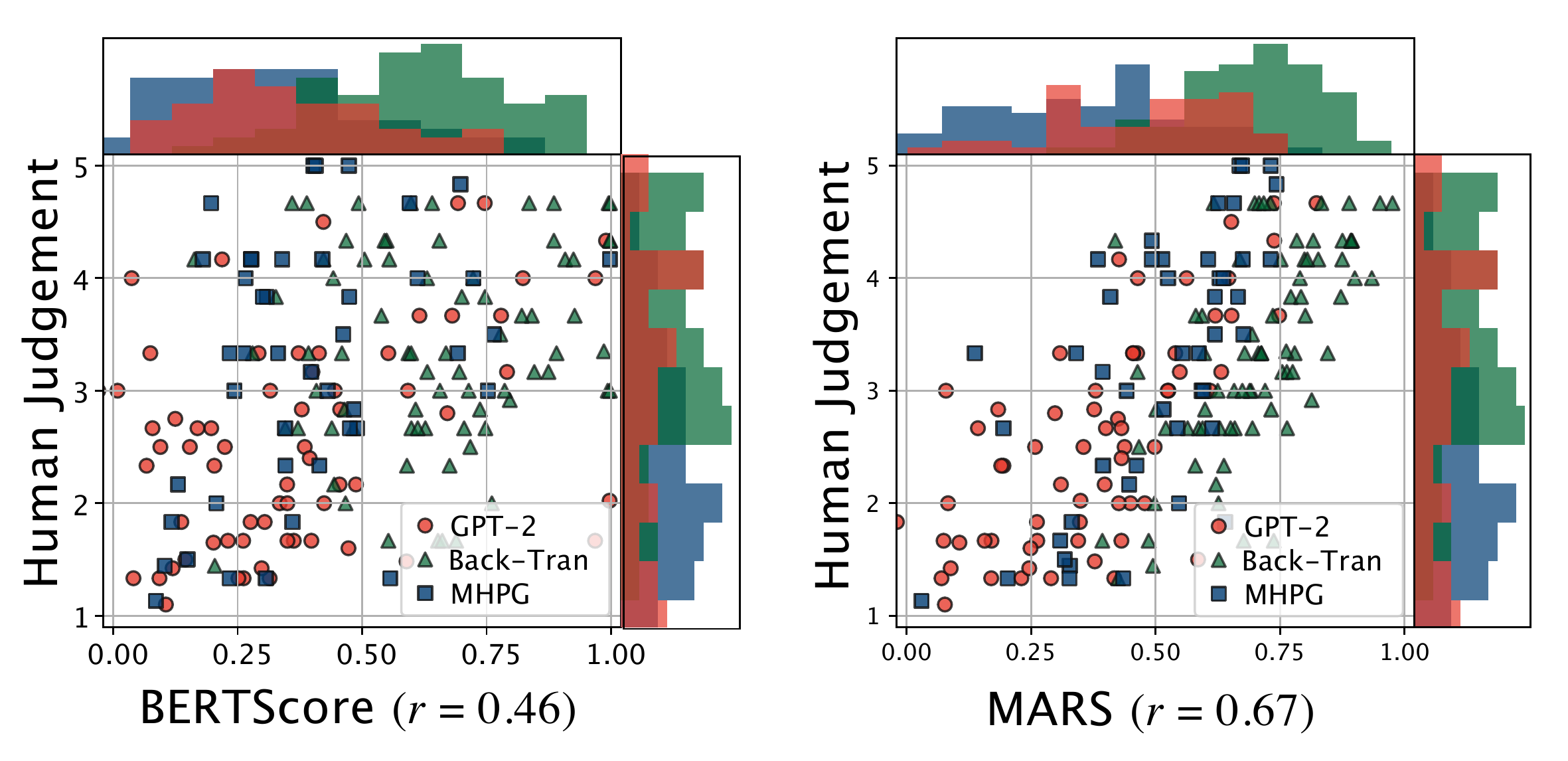}
   \vspace{-0.3in}
    \caption{Correlation between BERTScore (left) and MARS (right) with human judgements for MOCHA QA. The $x$-axis is the automated metric score and $y$-axis is the human judgement. \dpedit{Points in different colors represent generation outputs of three NLG systems: GPT-2 (red circles), Back-Translation (green triangles), and MHPG (blue squares)}.
    }
    \vspace{-0.2in}
   \label{fig:correlation}
 \end{figure}

To visualize the correlation between automated metrics and human judgements, we consider the MOCHA QA task as an example and plot the correlations of BERTScore (left) and MARS (right) with human judgements. As shown in Figure~\ref{fig:correlation}, compared with MARS, BERTScore has more candidates in the upper-left corner of the plot (i.e., low BERTScore but high human judgement). Many of these are generated by GPT-2 and MHPG, which, based on manual examination, tend to provide more details in the answer than the human reference. 
For instance, given a context about shopping, one question is \textit{``Did they need to buy any meat?''}. The human reference answer is simply \textit{``Yes, they did.''}, but GPT-2 returns \textit{``Yes, they bought chicken and a roast.''}, which is more detailed, even containing item names derived from the context.
Whereas BERTScore cannot evaluate such cases where the generated candidate is over-described with respect to the human reference, MARS uses augmented references enriched with information from the context to provide a fairer judgement.

\subsection{Is MARS robust?}
\label{subsec:robustness}

Good evaluation metrics ought to also be able to detect adversarial examples by assigning them lower scores than well-formed candidates.

As shown in Table~\ref{tab:robustness}, uni-gram matching BLEU-1 cannot detect reordered sequences, while ROUGE-L scores reordered sequence higher occasionally if token-swapping leads to more LCS. Sentence Mover's Similarity combines word and sentence embeddings and thus is more capable of recognizing reordered samples than MoverScore. Perplexity can detect reordered examples effectively, but is unable to detect retrieved sentences, as they are usually well-formed. 
MARS, on the other hand, has the best robustness against adversarial samples, possibly because multiple context-infused augmented references help MARS detect adversarial samples more reliably. \redit{We also study the effects of contextual embeddings we use in $\S$\ref{subsec:contextual_similarity}---when switching to GloVe embeddings~\cite{pennington-etal-2014-glove}, which are not contextual, MARS is less able to detect adversarial samples, especially reordered ones. The Naive method, which by default uses RoBERTa embedding, achieves comparable robustness as MARS but its task-level correlations with humans  (\textit{ref.}) are generally lower than MARS, potentially because its fixed-length \textit{cloze} generation limits the diversity of augmented references.}

\begin{table}[!ht]
\centering
\resizebox{0.49\textwidth}{!}{%
\begin{tabular}{@{}lccccc@{}}
\toprule
\multicolumn{6}{c}{\textbf{ROC Story Generation}}                                                         \\ \midrule
\textbf{$\{\lambda\}_{\textrm{max}}$} & 0\% (\textit{ref.}) & 20\%  & 40\%              & \dpboxblue{60\%}              & 80\%              \\ \midrule
\textbf{Pearson's $r$}              & 0.411          & 0.432 & 0.444   & \underline{0.459} & 0.452 \\
\textbf{Avg. $\sigma$}        &   -        & 0.027    &    0.046          &    0.055          & 0.059             \\ \midrule
\multicolumn{6}{c}{\textbf{Newsroom Summarization}}                                                          \\ \midrule
\textbf{$\{\lambda\}_\textrm{max}$} & 0\% (\textit{ref.}) & 20\%  & 40\%              & \dpboxblue{60\%}              & 80\%              \\ \midrule
\textbf{Pearson's $r$}              & 0.395          & 0.407 & 0.416 & \underline{0.423} & 0.411 \\
\textbf{Avg. $\sigma$}        &     -       & 0.061 & 0.062             & 0.063             & 0.068             \\ \midrule
\multicolumn{6}{c}{\textbf{MOCHA Question Answering}}                                                         \\ \midrule
\textbf{$\{\lambda\}_{\textrm{max}}$} & 0\% (\textit{ref.}) & \dpboxblue{20\%}  & 40\%              & 60\%              & 80\%              \\ \midrule
\textbf{Pearson's $r$}              & 0.658          & \underline{0.667} & 0.649 & 0.603 & 0.584 \\
\textbf{Avg. $\sigma$}        & -          & 0.074 & 0.104             & 0.117             & 0.125             \\ \bottomrule
\end{tabular}%
}
\caption{
Evaluating correlation with human judgements for various max masking ratios ($\lambda_{\textrm{max}}$) used in MARS.
0\% masking (\textit{ref.}) means only the human reference was used to score candidates.
We also show the averaged standard deviation of the cosine similarities between the candidate and augmented references across all samples.
}
\vspace{-0.2in}
\label{tab:best_param}
\end{table}
\subsection{Choosing Masking Ratios for MARS}

\label{subsec:best_masking}
The masking ratios for MARS are set using the hyperparameter $\{\lambda\}_{\textrm{max}}$, which corresponds to MARS using masking ratios from 0\% to $\{\lambda\}_{\textrm{max}}$ in increments of 20\%, e.g., $\{\lambda\}_{\textrm{max}}=40\%$ indicates $\lambda \in \{0\%, 20\%, 40\%\}$. 
In preliminary experiments, we observed that $\{\lambda\}_{\textrm{max}}$ varied for different datasets.
Thus, for our three generation tasks, we evaluate MARS performance given different $\{\lambda\}_\textrm{max}$, as shown in Table~\ref{tab:best_param}.
We find that tasks that were more open-ended (low $\Omega$; e.g., story generation) benefited from higher $\{\lambda\}_\textrm{max}$, which created a more diverse set of augmented references, whereas tasks that were less open-ended (high $\Omega$; e.g., QA) worked better with lower $\{\lambda\}_\textrm{max}$, which kept the augmented references more similar to the original.

\dpedit{

\subsection{Error Analysis}

We analyzed cases where MARS score substantially differed from human judgements. From test set outputs, we found that errors could often be categorized into one of three types (shown in Table~\ref{tab:error_analysis}): (1) \textbf{Out of Vocabulary} errors, often induced by unknown tokens in the candidates, (2) \textbf{Confusion} errors, where candidates are simply copied from context, and (3) \textbf{Inference} errors, where the candidates are further inferences of the context based on commonsense knowledge. In these cases, human annotators tended to assign higher scores, whereas, MARS over-penalized them.

\begingroup
\setlength{\tabcolsep}{3pt}
\begin{table}[]
\centering
\resizebox{0.47\textwidth}{!}{%
\begin{tabular}{@{}ll@{}}
\toprule
\textbf{Error} &
  \textbf{Example} \\ \midrule
  \begin{tabular}[c]{@{}l@{}}\textbf{OOV}\\ \textit{(ROC)}\end{tabular} &
  \begin{tabular}[c]{@{}l@{}}\textbf{Context}: ...waltz dance at wedding...\\ \textbf{Gold}: All the guests gasped \\ when they saw the couples' skill!\\ \textbf{Candidate}: All the guests gasped \\ when they saw the UNK UNK\\ \textbf{Human}: 0.392 \ \ \textbf{MARS}: 0.198\end{tabular} \\
 &
   \\
\begin{tabular}[c]{@{}l@{}}\textbf{Confusion}\\ \textit{(Newsroom)}\end{tabular} &
  \begin{tabular}[c]{@{}l@{}}\textbf{Context}: ...bidding on a neighborhood...\\ \textbf{Gold}: A neighborhood named \\ for its former orchards inspires loyalty \\ and bidding wars.\\ \textbf{Candidate}: Living there cherrydale lies \\ north of interstate... (a sentence extracted\\  from Context)\\ \textbf{Human}: 0.700 \ \ \textbf{MARS}: 0.399\end{tabular} \\
 &
   \\
   
\begin{tabular}[c]{@{}l@{}}\textbf{Inference}\\ \textit{(MOCHA)}\end{tabular} &
  \begin{tabular}[c]{@{}l@{}}\textbf{Context}: ...washing cloths... \\ \textbf{Q}: Why did they do the laundry?\\ \textbf{Gold}: To clean their clothes\\ \textbf{Candidate}: Because they were dirty.\\ \textbf{Human}: 0.400  \ \  \textbf{MARS}: 0.083\end{tabular} \\
   
  \bottomrule
\end{tabular}%
}
\caption{Error analysis of MARS. We investigated three typical types of errors within the samples which received large differences between the MARS score and human ratings. \textbf{Gold}: human written references.}
\label{tab:error_analysis}
\end{table}
\endgroup

\section{Human Judgement}

\begin{table*}[!ht]
\centering
\resizebox{0.85\textwidth}{!}{%
\begin{tabular}{@{}llccccccccc@{}}
\toprule
\multicolumn{2}{l}{\multirow{2}{*}{}} & \multicolumn{3}{c}{\cellcolor{RBRed}\textbf{ROC}} & \multicolumn{3}{c}{\cellcolor{RBGreen}\textbf{Newsroom}} & \multicolumn{3}{c}{\cellcolor{RBBlue}\textbf{MOCHA}} \\ \cmidrule(l){3-11} 
\multicolumn{2}{l}{}                               & Ori. & Naive  & MARS  & Ori. & Naive  & \multicolumn{1}{l}{MARS} & Ori. & Naive  & MARS   \\ \midrule
\multirow{2}{*}{\textbf{Relevance}}   & Mean       & 4.95 & 4.81   & 5.07  & 4.62 & 4.50   & 4.61                     & 5.16 & 4.61   & 4.97   \\
                                      & \textit{p} & -    & .00*  & .04*  & -    & .05   & .95                     & -    & .00*  & .10  \\ \cmidrule(r){1-2}
\multirow{2}{*}{\textbf{Readability}} & Mean       & 5.67 & 5.53   & 5.40  & 4.54 & 4.31   & 4.59                     & 5.41 & 5.23   & 5.33   \\
                                      & \textit{p} & -    & .11 & .05 & -    & .12 & .41                    & -    & .16 & .29 \\ \cmidrule(r){1-2}
\multirow{2}{*}{\textbf{Overall}}     & Mean       & 5.69 & 5.31   & 5.42  & 4.87 & 4.57   & 4.75                     & 4.62 & 4.44   & 4.68   \\
                                      & \textit{p} & -    & .12 & .30 & -    & .10 & .22                     & -    & .07  & .10   \\ \bottomrule
\end{tabular}%
}
    \caption{Human evaluation results on \textbf{Relevance} (to context), \textbf{Readability}, and \textbf{Overall} quality of MARS and Naive augmentation method. All results are compared with the original human reference (Ori.). Text was scored on a scale from 1-7. $p$ value describes the significance of difference. (* corresponds to $p<0.05$, ** to $p<0.01$
and *** to $p<0.001$.)}
\label{tab:human_eval}
\end{table*}

We conducted human evaluation on Amazon Mechanical Turk (MTurk) to further study the quality of MARS augmentation. In total 150 participants were randomly assigned to evaluate the three tasks. Participants (61.3\% male and 38.7\% female) were all from the United States and above 18 years old, with an average age of 34.7 years old. Each participant was paid 75 cents for completing 14 questions in each questionnaire (average completion time per questionnaire was about 5.11 minutes).

\paragraph{Results} We conducted paired sample $t$-tests to examine how much the augmentation samples resemble the original human references regarding relevance to context and readability. As shown in Table~\ref{tab:human_eval}, in terms of relevance to context, MARS had no statistically significant difference compared with original human references in Newsroom and MOCHA datasets, but was rated as even more relevant to the generation context than the human reference in the ROC dataset (MARS Mean = 5.07 $>$ Human Ref. Mean = 4.95), possibly because reinforced self-planning guided the augmentation to be more related to the context. In terms of readability, both MARS and Naive were rated lower than the original but not significantly; we take this as a compromise of \textit{cloze} style augmentation. No statistically significant differences were seen between the original and MARS augmentation in overall ratings across the three tasks. These results further confirm that augmented examples from MARS are of similar quality to the original human references.

}

\section{Related Metrics}
\label{sec:related}

\paragraph{Unsupervised Metrics.} 

In addition to the metrics we directly compared with previously, other unsupervised metrics have also been proposed.
TER~\cite{snover2006study}, CharacTer~\cite{wang-etal-2016-character}, and chrF~\cite{popovic-2017-chrf} focus on character-level overlaps instead of $n$-gram matching. Similar to BERTScore, YiSi~\cite{lo-2019-yisi} and BERTr~\cite{mathur-etal-2019-putting} leverage pre-trained contextual embeddings to better capture similarity.
$\Delta$BLEU~\cite{galley-etal-2015-deltableu} adds human annotated sentences as negative references. \citet{bawden-etal-2020-study} find the gain from multiple references can be limited by inherent weaknesses in BLEU. 
We considered lessons from many of the above works while designing MARS.

\paragraph{Learned Metrics.} Compared with unsupervised metrics, learned metrics collect human supervisions~\cite{freitag-etal-2020-human,chaganty-etal-2018-price} or train on specially prepared data of a certain domain~\cite{sellam-etal-2020-bleurt,rei2020comet}.
Other approaches train on related tasks and use these models as metrics for the original task \cite{goodrich2019assessing, eyal-etal-2019-question}. Whereas learned metrics may have limited applicability on tasks where no such resources are available, MARS fully exploits the few-shot learning abilities of off-the-shelf LMs and therefore does not require additional training.

\paragraph{Task-specific Metrics.} Finally, many metrics have been proposed for task-specific evaluation, such as  LEIC~\cite{cui2018learning} and CIDEr~\cite{vedantam2015cider} for image captioning, PARENT~\cite{dhingra-etal-2019-handling} for table-to-text, and EASSE~\cite{alva-manchego-etal-2019-easse} for sentence simplification. MARS, with some modifications, can potentially be extended to these tasks.

\section{Limitations}
MARS can be limited by the LM that it uses---for instance, the total length of context + reference/candidate is limited by the max sequence length of the LM used. 
Additionally, our work has focused on English, and MARS may require non-trivial modifications to handle cases where the context and reference/candidate are in different languages, such as machine translation.
Future work, could potentially extend MARS to these scenarios using multi-lingual sequence-to-sequence models such as multilingual-T5 \cite{xue2020mt5}. 
We also analyzed errors and found that MARS sometimes under-scores candidates that contained unknown tokens or were copied directly from the context (see Appendix C for examples and further analysis).

\section{Conclusion}

We have proposed MARS, a context-aware and easy-to-deploy NLG metric built upon an off-the-shelf language model (GPT-2). On three contextual NLG tasks, we show that MARS better correlates with human judgements compared with seven other unsupervised metrics. Requiring neither costly human supervision nor additional training, MARS can be applied to a broad range of NLG tasks.

\dpedit{
\section*{Ethical Considerations}
The goal of MARS is to aid the evaluation of NLG models, and hence we draw attention to several ethical considerations. First, the augmented references of MARS can be affected by certain biases from the LM it is based on (e.g., GPT-2)~\citep{liu2021mitigating}, though those biases may be partially mitigated by the relatively narrow scope of \textit{cloze} completion and by generations being guided by given context and human references. Second, MARS facilitates evaluation and therefore development of NLG models, for which a major ethical consideration is that they can mimic target properties in training data that are undesirable. This is especially true of models trained on non-contemporary data that does not represent current norms and practices. These biases can lead to ethical concerns if users or deployers of models are not aware of these issues or do not account for them. More generally, NLG models can also be used in malicious ways such as to generate fake news or spam, which we strongly discourage. Finally, our experiments and analysis are done in English, and therefore we do not claim that our findings will generalize across all languages, although our framework has potential to be extended to other languages with necessary modifications.
}

\bibliographystyle{acl_natbib}
\bibliography{acl2021}

\begin{thebibliography}{64}
\expandafter\ifx\csname natexlab\endcsname\relax\def\natexlab#1{#1}\fi

\bibitem[{Alva-Manchego et~al.(2019)Alva-Manchego, Martin, Scarton, and
  Specia}]{alva-manchego-etal-2019-easse}
Fernando Alva-Manchego, Louis Martin, Carolina Scarton, and Lucia Specia. 2019.
\newblock \href {https://doi.org/10.18653/v1/D19-3009} {{EASSE}: Easier
  automatic sentence simplification evaluation}.
\newblock In \emph{Proceedings of the 2019 Conference on Empirical Methods in
  Natural Language Processing and the 9th International Joint Conference on
  Natural Language Processing (EMNLP-IJCNLP): System Demonstrations}, pages
  49--54, Hong Kong, China. Association for Computational Linguistics.

\bibitem[{Bauer et~al.(2018)Bauer, Wang, and Bansal}]{bauer2018commonsense}
Lisa Bauer, Yicheng Wang, and Mohit Bansal. 2018.
\newblock \href {https://www.aclweb.org/anthology/D18-1454.pdf} {Commonsense
  for generative multi-hop question answering tasks}.
\newblock In \emph{Proceedings of the 2018 Conference on Empirical Methods in
  Natural Language Processing}, pages 4220--4230.

\bibitem[{Bawden et~al.(2020)Bawden, Zhang, Yankovskaya, T{\"a}ttar, and
  Post}]{bawden-etal-2020-study}
Rachel Bawden, Biao Zhang, Lisa Yankovskaya, Andre T{\"a}ttar, and Matt Post.
  2020.
\newblock \href {https://doi.org/10.18653/v1/2020.findings-emnlp.82} {A study
  in improving {BLEU} reference coverage with diverse automatic paraphrasing}.
\newblock In \emph{Findings of the Association for Computational Linguistics:
  EMNLP 2020}, pages 918--932, Online. Association for Computational
  Linguistics.

\bibitem[{Brown et~al.(1992)Brown, Della~Pietra, Della~Pietra, Lai, and
  Mercer}]{brown1992estimate}
Peter~F Brown, Stephen~A Della~Pietra, Vincent~J Della~Pietra, Jennifer~C Lai,
  and Robert~L Mercer. 1992.
\newblock \href {https://www.aclweb.org/anthology/J92-1002.pdf} {An estimate of
  an upper bound for the entropy of english}.
\newblock \emph{Computational Linguistics}, 18(1):31--40.

\bibitem[{Chaganty et~al.(2018)Chaganty, Mussmann, and
  Liang}]{chaganty-etal-2018-price}
Arun Chaganty, Stephen Mussmann, and Percy Liang. 2018.
\newblock \href {https://doi.org/10.18653/v1/P18-1060} {The price of debiasing
  automatic metrics in natural language evalaution}.
\newblock In \emph{Proceedings of the 56th Annual Meeting of the Association
  for Computational Linguistics (Volume 1: Long Papers)}, pages 643--653,
  Melbourne, Australia. Association for Computational Linguistics.

\bibitem[{Chen et~al.(2020)Chen, Stanovsky, Singh, and Gardner}]{chen2020mocha}
Anthony Chen, Gabriel Stanovsky, Sameer Singh, and Matt Gardner. 2020.
\newblock \href {https://www.aclweb.org/anthology/2020.emnlp-main.528.pdf}
  {Mocha: A dataset for training and evaluating generative reading
  comprehension metrics}.
\newblock In \emph{Proceedings of the 2020 Conference on Empirical Methods in
  Natural Language Processing (EMNLP)}, pages 6521--6532.

\bibitem[{Clark et~al.(2019)Clark, Celikyilmaz, and
  Smith}]{clark-etal-2019-sentence}
Elizabeth Clark, Asli Celikyilmaz, and Noah~A. Smith. 2019.
\newblock \href {https://doi.org/10.18653/v1/P19-1264} {Sentence mover{'}s
  similarity: Automatic evaluation for multi-sentence texts}.
\newblock In \emph{Proceedings of the 57th Annual Meeting of the Association
  for Computational Linguistics}, pages 2748--2760, Florence, Italy.
  Association for Computational Linguistics.

\bibitem[{Cui et~al.(2018)Cui, Yang, Veit, Huang, and
  Belongie}]{cui2018learning}
Yin Cui, Guandao Yang, Andreas Veit, Xun Huang, and Serge Belongie. 2018.
\newblock \href
  {https://openaccess.thecvf.com/content_cvpr_2018/papers/Cui_Learning_to_Evaluate_CVPR_2018_paper.pdf}
  {Learning to evaluate image captioning}.
\newblock In \emph{Proceedings of the IEEE conference on computer vision and
  pattern recognition}, pages 5804--5812.

\bibitem[{Devlin et~al.(2019)Devlin, Chang, Lee, and
  Toutanova}]{devlin2019bert}
Jacob Devlin, Ming-Wei Chang, Kenton Lee, and Kristina Toutanova. 2019.
\newblock \href {https://www.aclweb.org/anthology/N19-1423.pdf} {Bert:
  Pre-training of deep bidirectional transformers for language understanding}.
\newblock In \emph{Proceedings of the 2019 Conference of the North American
  Chapter of the Association for Computational Linguistics: Human Language
  Technologies, Volume 1 (Long and Short Papers)}, pages 4171--4186.

\bibitem[{Dhingra et~al.(2019)Dhingra, Faruqui, Parikh, Chang, Das, and
  Cohen}]{dhingra-etal-2019-handling}
Bhuwan Dhingra, Manaal Faruqui, Ankur Parikh, Ming-Wei Chang, Dipanjan Das, and
  William Cohen. 2019.
\newblock \href {https://doi.org/10.18653/v1/P19-1483} {Handling divergent
  reference texts when evaluating table-to-text generation}.
\newblock In \emph{Proceedings of the 57th Annual Meeting of the Association
  for Computational Linguistics}, pages 4884--4895, Florence, Italy.
  Association for Computational Linguistics.

\bibitem[{Donahue et~al.(2020)Donahue, Lee, and
  Liang}]{donahue-etal-2020-enabling}
Chris Donahue, Mina Lee, and Percy Liang. 2020.
\newblock \href {https://doi.org/10.18653/v1/2020.acl-main.225} {Enabling
  language models to fill in the blanks}.
\newblock In \emph{Proceedings of the 58th Annual Meeting of the Association
  for Computational Linguistics}, pages 2492--2501, Online. Association for
  Computational Linguistics.

\bibitem[{Durmus et~al.(2020)Durmus, He, and Diab}]{durmus2020feqa}
Esin Durmus, He~He, and Mona Diab. 2020.
\newblock \href {https://doi.org/10.18653/v1/2020.acl-main.454} {{FEQA}: A
  question answering evaluation framework for faithfulness assessment in
  abstractive summarization}.
\newblock In \emph{Proceedings of the 58th Annual Meeting of the Association
  for Computational Linguistics}, pages 5055--5070, Online. Association for
  Computational Linguistics.

\bibitem[{Eyal et~al.(2019)Eyal, Baumel, and Elhadad}]{eyal-etal-2019-question}
Matan Eyal, Tal Baumel, and Michael Elhadad. 2019.
\newblock \href {https://doi.org/10.18653/v1/N19-1395} {Question answering as
  an automatic evaluation metric for news article summarization}.
\newblock In \emph{Proceedings of the 2019 Conference of the North {A}merican
  Chapter of the Association for Computational Linguistics: Human Language
  Technologies, Volume 1 (Long and Short Papers)}, pages 3938--3948,
  Minneapolis, Minnesota. Association for Computational Linguistics.

\bibitem[{Freitag et~al.(2020{\natexlab{a}})Freitag, Foster, Grangier, and
  Cherry}]{freitag-etal-2020-human}
Markus Freitag, George Foster, David Grangier, and Colin Cherry.
  2020{\natexlab{a}}.
\newblock \href {https://www.aclweb.org/anthology/2020.wmt-1.140}
  {Human-paraphrased references improve neural machine translation}.
\newblock In \emph{Proceedings of the Fifth Conference on Machine Translation},
  pages 1183--1192, Online. Association for Computational Linguistics.

\bibitem[{Freitag et~al.(2020{\natexlab{b}})Freitag, Grangier, and
  Caswell}]{freitag-etal-2020-bleu}
Markus Freitag, David Grangier, and Isaac Caswell. 2020{\natexlab{b}}.
\newblock \href {https://doi.org/10.18653/v1/2020.emnlp-main.5} {{BLEU} might
  be guilty but references are not innocent}.
\newblock In \emph{Proceedings of the 2020 Conference on Empirical Methods in
  Natural Language Processing (EMNLP)}, pages 61--71, Online. Association for
  Computational Linguistics.

\bibitem[{Galley et~al.(2015)Galley, Brockett, Sordoni, Ji, Auli, Quirk,
  Mitchell, Gao, and Dolan}]{galley-etal-2015-deltableu}
Michel Galley, Chris Brockett, Alessandro Sordoni, Yangfeng Ji, Michael Auli,
  Chris Quirk, Margaret Mitchell, Jianfeng Gao, and Bill Dolan. 2015.
\newblock \href {https://doi.org/10.3115/v1/P15-2073} {delta{BLEU}: A
  discriminative metric for generation tasks with intrinsically diverse
  targets}.
\newblock In \emph{Proceedings of the 53rd Annual Meeting of the Association
  for Computational Linguistics and the 7th International Joint Conference on
  Natural Language Processing (Volume 2: Short Papers)}, pages 445--450,
  Beijing, China. Association for Computational Linguistics.

\bibitem[{Gong et~al.(2019)Gong, Bhat, Wu, Xiong, and
  Hwu}]{gong-etal-2019-reinforcement}
Hongyu Gong, Suma Bhat, Lingfei Wu, JinJun Xiong, and Wen-mei Hwu. 2019.
\newblock \href {https://doi.org/10.18653/v1/N19-1320} {Reinforcement learning
  based text style transfer without parallel training corpus}.
\newblock In \emph{Proceedings of the 2019 Conference of the North {A}merican
  Chapter of the Association for Computational Linguistics: Human Language
  Technologies, Volume 1 (Long and Short Papers)}, pages 3168--3180,
  Minneapolis, Minnesota. Association for Computational Linguistics.

\bibitem[{Goodrich et~al.(2019)Goodrich, Saleh, Liu, and
  Rao}]{goodrich2019assessing}
Ben Goodrich, Mohammad~Ahmad Saleh, Peter Liu, and Vinay Rao. 2019.
\newblock \href {https://dl.acm.org/doi/pdf/10.1145/3292500.3330955} {Assessing
  the factual accuracy of text generation}.

\bibitem[{Grusky et~al.(2018)Grusky, Naaman, and
  Artzi}]{grusky-etal-2018-newsroom}
Max Grusky, Mor Naaman, and Yoav Artzi. 2018.
\newblock \href {https://doi.org/10.18653/v1/N18-1065} {{N}ewsroom: A dataset
  of 1.3 million summaries with diverse extractive strategies}.
\newblock In \emph{Proceedings of the 2018 Conference of the North {A}merican
  Chapter of the Association for Computational Linguistics: Human Language
  Technologies, Volume 1 (Long Papers)}, pages 708--719, New Orleans,
  Louisiana. Association for Computational Linguistics.

\bibitem[{Gupta et~al.(2019)Gupta, Mehri, Zhao, Pavel, Eskenazi, and
  Bigham}]{gupta2019investigating}
Prakhar Gupta, Shikib Mehri, Tiancheng Zhao, Amy Pavel, Maxine Eskenazi, and
  Jeffrey~P Bigham. 2019.
\newblock \href {https://www.aclweb.org/anthology/W19-5944.pdf} {Investigating
  evaluation of open-domain dialogue systems with human generated multiple
  references}.
\newblock In \emph{Proceedings of the 20th Annual SIGdial Meeting on Discourse
  and Dialogue}, pages 379--391.

\bibitem[{Hashimoto et~al.(2019)Hashimoto, Zhang, and
  Liang}]{hashimoto-etal-2019-unifying}
Tatsunori Hashimoto, Hugh Zhang, and Percy Liang. 2019.
\newblock \href {https://doi.org/10.18653/v1/N19-1169} {Unifying human and
  statistical evaluation for natural language generation}.
\newblock In \emph{Proceedings of the 2019 Conference of the North {A}merican
  Chapter of the Association for Computational Linguistics: Human Language
  Technologies, Volume 1 (Long and Short Papers)}, pages 1689--1701,
  Minneapolis, Minnesota. Association for Computational Linguistics.

\bibitem[{Keskar et~al.(2019)Keskar, McCann, Varshney, Xiong, and
  Socher}]{keskar2019ctrl}
Nitish~Shirish Keskar, Bryan McCann, Lav~R Varshney, Caiming Xiong, and Richard
  Socher. 2019.
\newblock \href {https://arxiv.org/pdf/1909.05858.pdf} {Ctrl: A conditional
  transformer language model for controllable generation}.
\newblock \emph{arXiv preprint arXiv:1909.05858}.

\bibitem[{Ko{\v{c}}isk{\`y} et~al.(2018)Ko{\v{c}}isk{\`y}, Schwarz, Blunsom,
  Dyer, Hermann, Melis, and Grefenstette}]{kovcisky2018narrativeqa}
Tom{\'a}{\v{s}} Ko{\v{c}}isk{\`y}, Jonathan Schwarz, Phil Blunsom, Chris Dyer,
  Karl~Moritz Hermann, G{\'a}bor Melis, and Edward Grefenstette. 2018.
\newblock \href {https://www.aclweb.org/anthology/Q18-1023.pdf} {The
  narrativeqa reading comprehension challenge}.
\newblock \emph{Transactions of the Association for Computational Linguistics},
  6:317--328.

\bibitem[{Lavie and Agarwal(2007)}]{lavie-agarwal-2007-meteor}
Alon Lavie and Abhaya Agarwal. 2007.
\newblock \href {https://www.aclweb.org/anthology/W07-0734} {{METEOR}: An
  automatic metric for {MT} evaluation with high levels of correlation with
  human judgments}.
\newblock In \emph{Proceedings of the Second Workshop on Statistical Machine
  Translation}, pages 228--231, Prague, Czech Republic. Association for
  Computational Linguistics.

\bibitem[{Lin(2004)}]{lin2004rouge}
Chin-Yew Lin. 2004.
\newblock \href {https://www.aclweb.org/anthology/W04-1013.pdf} {Rouge: A
  package for automatic evaluation of summaries}.
\newblock In \emph{Text summarization branches out}, pages 74--81.

\bibitem[{Lin and Och(2004)}]{lin-och-2004-automatic}
Chin-Yew Lin and Franz~Josef Och. 2004.
\newblock \href {https://doi.org/10.3115/1218955.1219032} {Automatic evaluation
  of machine translation quality using longest common subsequence and
  skip-bigram statistics}.
\newblock In \emph{Proceedings of the 42nd Annual Meeting of the Association
  for Computational Linguistics ({ACL}-04)}, pages 605--612, Barcelona, Spain.

\bibitem[{Liu et~al.(2021)Liu, Jia, Wei, Xu, Wang, and
  Vosoughi}]{liu2021mitigating}
Ruibo Liu, Chenyan Jia, Jason Wei, Guangxuan Xu, Lili Wang, and Soroush
  Vosoughi. 2021.
\newblock Mitigating political bias in language models through reinforced
  calibration.
\newblock In \emph{Proceedings of the AAAI Conference on Artificial
  Intelligence}.

\bibitem[{Liu et~al.(2019)Liu, Ott, Goyal, Du, Joshi, Chen, Levy, Lewis,
  Zettlemoyer, and Stoyanov}]{liu2019roberta}
Yinhan Liu, Myle Ott, Naman Goyal, Jingfei Du, Mandar Joshi, Danqi Chen, Omer
  Levy, Mike Lewis, Luke Zettlemoyer, and Veselin Stoyanov. 2019.
\newblock \href {https://arxiv.org/pdf/1907.11692.pdf} {Roberta: A robustly
  optimized bert pretraining approach}.
\newblock \emph{arXiv preprint arXiv:1907.11692}.

\bibitem[{Lo(2019)}]{lo-2019-yisi}
Chi-kiu Lo. 2019.
\newblock \href {https://doi.org/10.18653/v1/W19-5358} {{Y}i{S}i - a unified
  semantic {MT} quality evaluation and estimation metric for languages with
  different levels of available resources}.
\newblock In \emph{Proceedings of the Fourth Conference on Machine Translation
  (Volume 2: Shared Task Papers, Day 1)}, pages 507--513, Florence, Italy.
  Association for Computational Linguistics.

\bibitem[{Mathur et~al.(2019)Mathur, Baldwin, and
  Cohn}]{mathur-etal-2019-putting}
Nitika Mathur, Timothy Baldwin, and Trevor Cohn. 2019.
\newblock \href {https://doi.org/10.18653/v1/P19-1269} {Putting evaluation in
  context: Contextual embeddings improve machine translation evaluation}.
\newblock In \emph{Proceedings of the 57th Annual Meeting of the Association
  for Computational Linguistics}, pages 2799--2808, Florence, Italy.
  Association for Computational Linguistics.

\bibitem[{McCann et~al.(2017)McCann, Bradbury, Xiong, and
  Socher}]{mccann2017learned}
Bryan McCann, James Bradbury, Caiming Xiong, and Richard Socher. 2017.
\newblock \href
  {https://papers.nips.cc/paper/2017/file/20c86a628232a67e7bd46f76fba7ce12-Paper.pdf}
  {Learned in translation: Contextualized word vectors}.
\newblock In \emph{Advances in Neural Information Processing Systems}, pages
  6294--6305.

\bibitem[{Mehri and Eskenazi(2020)}]{mehri-eskenazi-2020-usr}
Shikib Mehri and Maxine Eskenazi. 2020.
\newblock \href {https://doi.org/10.18653/v1/2020.acl-main.64} {{USR}: An
  unsupervised and reference free evaluation metric for dialog generation}.
\newblock In \emph{Proceedings of the 58th Annual Meeting of the Association
  for Computational Linguistics}, pages 681--707, Online. Association for
  Computational Linguistics.

\bibitem[{Miao et~al.(2019)Miao, Zhou, Mou, Yan, and Li}]{miao2019cgmh}
Ning Miao, Hao Zhou, Lili Mou, Rui Yan, and Lei Li. 2019.
\newblock \href {https://ojs.aaai.org/index.php/AAAI/article/view/4659/4537}
  {Cgmh: Constrained sentence generation by metropolis-hastings sampling}.
\newblock In \emph{Proceedings of the AAAI Conference on Artificial
  Intelligence}, volume~33, pages 6834--6842.

\bibitem[{Mostafazadeh et~al.(2016)Mostafazadeh, Chambers, He, Parikh, Batra,
  Vanderwende, Kohli, and Allen}]{mostafazadeh-etal-2016-corpus}
Nasrin Mostafazadeh, Nathanael Chambers, Xiaodong He, Devi Parikh, Dhruv Batra,
  Lucy Vanderwende, Pushmeet Kohli, and James Allen. 2016.
\newblock \href {https://doi.org/10.18653/v1/N16-1098} {A corpus and cloze
  evaluation for deeper understanding of commonsense stories}.
\newblock In \emph{Proceedings of the 2016 Conference of the North {A}merican
  Chapter of the Association for Computational Linguistics: Human Language
  Technologies}, pages 839--849, San Diego, California. Association for
  Computational Linguistics.

\bibitem[{Munos et~al.(2016)Munos, Stepleton, Harutyunyan, and
  Bellemare}]{munos2016safe}
R{\'e}mi Munos, Tom Stepleton, Anna Harutyunyan, and Marc Bellemare. 2016.
\newblock \href
  {http://papers.nips.cc/paper/6538-safe-and-efficient-off-policy-reinforcement-learning.pdf}
  {Safe and efficient off-policy reinforcement learning}.
\newblock In \emph{Advances in Neural Information Processing Systems}, pages
  1054--1062.

\bibitem[{Nema and Khapra(2018)}]{nema-khapra-2018-towards}
Preksha Nema and Mitesh~M. Khapra. 2018.
\newblock \href {https://doi.org/10.18653/v1/D18-1429} {Towards a better metric
  for evaluating question generation systems}.
\newblock In \emph{Proceedings of the 2018 Conference on Empirical Methods in
  Natural Language Processing}, pages 3950--3959, Brussels, Belgium.
  Association for Computational Linguistics.

\bibitem[{Novikova et~al.(2017)Novikova, Du{\v{s}}ek, Cercas~Curry, and
  Rieser}]{novikova-etal-2017-need}
Jekaterina Novikova, Ond{\v{r}}ej Du{\v{s}}ek, Amanda Cercas~Curry, and Verena
  Rieser. 2017.
\newblock \href {https://doi.org/10.18653/v1/D17-1238} {Why we need new
  evaluation metrics for {NLG}}.
\newblock In \emph{Proceedings of the 2017 Conference on Empirical Methods in
  Natural Language Processing}, pages 2241--2252, Copenhagen, Denmark.
  Association for Computational Linguistics.

\bibitem[{Ostermann et~al.(2018)Ostermann, Modi, Roth, Thater, and
  Pinkal}]{ostermann2018mcscript}
Simon Ostermann, Ashutosh Modi, Michael Roth, Stefan Thater, and Manfred
  Pinkal. 2018.
\newblock \href {https://www.aclweb.org/anthology/L18-1564.pdf} {Mcscript: A
  novel dataset for assessing machine comprehension using script knowledge}.
\newblock In \emph{Proceedings of the Eleventh International Conference on
  Language Resources and Evaluation (LREC 2018)}.

\bibitem[{Page et~al.(1999)Page, Brin, Motwani, and
  Winograd}]{page1999pagerank}
Lawrence Page, Sergey Brin, Rajeev Motwani, and Terry Winograd. 1999.
\newblock \href {http://ilpubs.stanford.edu:8090/422/1/1999-66.pdf} {The
  pagerank citation ranking: Bringing order to the web.}
\newblock Technical report, Stanford InfoLab.

\bibitem[{Pang and He(2021)}]{pang2020text}
Richard~Yuanzhe Pang and He~He. 2021.
\newblock \href {arXiv preprint arXiv:2009.07839} {Text generation by learning
  from off-policy demonstrations}.
\newblock \emph{International Conference on Learning Representations (ICLR
  21')}.

\bibitem[{Papineni et~al.(2002)Papineni, Roukos, Ward, and
  Zhu}]{papineni2002bleu}
Kishore Papineni, Salim Roukos, Todd Ward, and Wei-Jing Zhu. 2002.
\newblock \href {https://www.aclweb.org/anthology/P02-1040.pdf} {Bleu: a method
  for automatic evaluation of machine translation}.
\newblock In \emph{Proceedings of the 40th annual meeting of the Association
  for Computational Linguistics}, pages 311--318.

\bibitem[{Pennington et~al.(2014)Pennington, Socher, and
  Manning}]{pennington-etal-2014-glove}
Jeffrey Pennington, Richard Socher, and Christopher Manning. 2014.
\newblock \href {https://doi.org/10.3115/v1/D14-1162} {{G}lo{V}e: Global
  vectors for word representation}.
\newblock In \emph{Proceedings of the 2014 Conference on Empirical Methods in
  Natural Language Processing ({EMNLP})}, pages 1532--1543, Doha, Qatar.
  Association for Computational Linguistics.

\bibitem[{Pisinger(1995)}]{pisinger1995algorithms}
David Pisinger. 1995.
\newblock \href
  {http://citeseerx.ist.psu.edu/viewdoc/download?doi=10.1.1.16.9780&rep=rep1&type=pdf}
  {Algorithms for knapsack problems}.

\bibitem[{Popovi{\'c}(2017)}]{popovic-2017-chrf}
Maja Popovi{\'c}. 2017.
\newblock \href {https://doi.org/10.18653/v1/W17-4770} {chr{F}++: words helping
  character n-grams}.
\newblock In \emph{Proceedings of the Second Conference on Machine
  Translation}, pages 612--618, Copenhagen, Denmark. Association for
  Computational Linguistics.

\bibitem[{Radford et~al.(2019)Radford, Wu, Child, Luan, Amodei, and
  Sutskever}]{radford2019language}
Alec Radford, Jeffrey Wu, Rewon Child, David Luan, Dario Amodei, and Ilya
  Sutskever. 2019.
\newblock \href
  {https://d4mucfpksywv.cloudfront.net/better-language-models/language-models.pdf}
  {Language models are unsupervised multitask learners}.
\newblock \emph{OpenAI Blog}, 1(8):9.

\bibitem[{Rei et~al.(2020)Rei, Stewart, Farinha, and Lavie}]{rei2020comet}
Ricardo Rei, Craig Stewart, Ana~C Farinha, and Alon Lavie. 2020.
\newblock \href {https://doi.org/10.18653/v1/2020.emnlp-main.213} {{COMET}: A
  neural framework for {MT} evaluation}.
\newblock In \emph{Proceedings of the 2020 Conference on Empirical Methods in
  Natural Language Processing (EMNLP)}, pages 2685--2702, Online. Association
  for Computational Linguistics.

\bibitem[{Reimers and Gurevych(2019)}]{reimers-2019-sentence-bert}
Nils Reimers and Iryna Gurevych. 2019.
\newblock \href {https://arxiv.org/abs/1908.10084} {Sentence-bert: Sentence
  embeddings using siamese bert-networks}.
\newblock In \emph{Proceedings of the 2019 Conference on Empirical Methods in
  Natural Language Processing}. Association for Computational Linguistics.

\bibitem[{Reimers and Gurevych(2020)}]{reimers-2020-multilingual-sentence-bert}
Nils Reimers and Iryna Gurevych. 2020.
\newblock \href {https://arxiv.org/abs/2004.09813} {Making monolingual sentence
  embeddings multilingual using knowledge distillation}.
\newblock In \emph{Proceedings of the 2020 Conference on Empirical Methods in
  Natural Language Processing}. Association for Computational Linguistics.

\bibitem[{Rush et~al.(2015)Rush, Chopra, and Weston}]{rush-etal-2015-neural}
Alexander~M. Rush, Sumit Chopra, and Jason Weston. 2015.
\newblock \href {https://doi.org/10.18653/v1/D15-1044} {A neural attention
  model for abstractive sentence summarization}.
\newblock In \emph{Proceedings of the 2015 Conference on Empirical Methods in
  Natural Language Processing}, pages 379--389, Lisbon, Portugal. Association
  for Computational Linguistics.

\bibitem[{See et~al.(2017)See, Liu, and Manning}]{see2017get}
Abigail See, Peter~J Liu, and Christopher~D Manning. 2017.
\newblock \href {https://www.aclweb.org/anthology/P17-1099.pdf} {Get to the
  point: Summarization with pointer-generator networks}.
\newblock In \emph{Proceedings of the 55th Annual Meeting of the Association
  for Computational Linguistics (Volume 1: Long Papers)}, pages 1073--1083.

\bibitem[{Sellam et~al.(2020)Sellam, Das, and Parikh}]{sellam-etal-2020-bleurt}
Thibault Sellam, Dipanjan Das, and Ankur Parikh. 2020.
\newblock \href {https://doi.org/10.18653/v1/2020.acl-main.704} {{BLEURT}:
  Learning robust metrics for text generation}.
\newblock In \emph{Proceedings of the 58th Annual Meeting of the Association
  for Computational Linguistics}, pages 7881--7892, Online. Association for
  Computational Linguistics.

\bibitem[{Sennrich et~al.(2016)Sennrich, Haddow, and
  Birch}]{sennrich2016improving}
Rico Sennrich, Barry Haddow, and Alexandra Birch. 2016.
\newblock \href {https://www.aclweb.org/anthology/P16-1009.pdf} {Improving
  neural machine translation models with monolingual data}.
\newblock In \emph{Proceedings of the 54th Annual Meeting of the Association
  for Computational Linguistics (Volume 1: Long Papers)}, pages 86--96.

\bibitem[{Shen et~al.(2020)Shen, Quach, Barzilay, and
  Jaakkola}]{shen-etal-2020-blank}
Tianxiao Shen, Victor Quach, Regina Barzilay, and Tommi Jaakkola. 2020.
\newblock \href {https://doi.org/10.18653/v1/2020.emnlp-main.420} {Blank
  language models}.
\newblock In \emph{Proceedings of the 2020 Conference on Empirical Methods in
  Natural Language Processing (EMNLP)}, pages 5186--5198, Online. Association
  for Computational Linguistics.

\bibitem[{Snover et~al.(2006)Snover, Dorr, Schwartz, Micciulla, and
  Makhoul}]{snover2006study}
Matthew Snover, Bonnie Dorr, Richard Schwartz, Linnea Micciulla, and John
  Makhoul. 2006.
\newblock \href {http://mt-archive.info/AMTA-2006-Snover.pdf} {A study of
  translation edit rate with targeted human annotation}.
\newblock In \emph{Proceedings of association for machine translation in the
  Americas}, volume 200. Cambridge, MA.

\bibitem[{Tao et~al.(2018)Tao, Mou, Zhao, and Yan}]{tao2018ruber}
Chongyang Tao, Lili Mou, Dongyan Zhao, and Rui Yan. 2018.
\newblock \href {https://ojs.aaai.org/index.php/AAAI/article/view/11321/11180}
  {Ruber: An unsupervised method for automatic evaluation of open-domain dialog
  systems}.
\newblock In \emph{Proceedings of the AAAI Conference on Artificial
  Intelligence}, volume~32.

\bibitem[{Taylor(1953)}]{taylor1953cloze}
Wilson~L Taylor. 1953.
\newblock “cloze procedure”: A new tool for measuring readability.
\newblock \emph{Journalism quarterly}, 30(4):415--433.

\bibitem[{Vedantam et~al.(2015)Vedantam, Lawrence~Zitnick, and
  Parikh}]{vedantam2015cider}
Ramakrishna Vedantam, C~Lawrence~Zitnick, and Devi Parikh. 2015.
\newblock \href
  {https://openaccess.thecvf.com/content_cvpr_2015/papers/Vedantam_CIDEr_Consensus-Based_Image_2015_CVPR_paper.pdf}
  {Cider: Consensus-based image description evaluation}.
\newblock In \emph{Proceedings of the IEEE conference on computer vision and
  pattern recognition}, pages 4566--4575.

\bibitem[{Wang et~al.(2016)Wang, Peter, Rosendahl, and
  Ney}]{wang-etal-2016-character}
Weiyue Wang, Jan-Thorsten Peter, Hendrik Rosendahl, and Hermann Ney. 2016.
\newblock \href {https://doi.org/10.18653/v1/W16-2342} {{C}harac{T}er:
  Translation edit rate on character level}.
\newblock In \emph{Proceedings of the First Conference on Machine Translation:
  Volume 2, Shared Task Papers}, pages 505--510, Berlin, Germany. Association
  for Computational Linguistics.

\bibitem[{Xue et~al.(2020)Xue, Constant, Roberts, Kale, Al-Rfou, Siddhant,
  Barua, and Raffel}]{xue2020mt5}
Linting Xue, Noah Constant, Adam Roberts, Mihir Kale, Rami Al-Rfou, Aditya
  Siddhant, Aditya Barua, and Colin Raffel. 2020.
\newblock mt5: A massively multilingual pre-trained text-to-text transformer.
\newblock \emph{arXiv preprint arXiv:2010.11934}.

\bibitem[{Yuma et~al.(2020)Yuma, Yoshinaga, and Toyoda}]{yuma-etal-2020-ubleu}
Tsuta Yuma, Naoki Yoshinaga, and Masashi Toyoda. 2020.
\newblock \href {https://doi.org/10.18653/v1/2020.acl-srw.27} {u{BLEU}:
  Uncertainty-aware automatic evaluation method for open-domain dialogue
  systems}.
\newblock In \emph{Proceedings of the 58th Annual Meeting of the Association
  for Computational Linguistics: Student Research Workshop}, pages 199--206,
  Online. Association for Computational Linguistics.

\bibitem[{Zhang et~al.(2019)Zhang, Kishore, Wu, Weinberger, and
  Artzi}]{zhang2019bertscore}
Tianyi Zhang, Varsha Kishore, Felix Wu, Kilian~Q Weinberger, and Yoav Artzi.
  2019.
\newblock \href {https://arxiv.org/pdf/1904.09675.pdf} {Bertscore: Evaluating
  text generation with bert}.

\bibitem[{Zhang et~al.(2020)Zhang, Wang, Li, Gan, Brockett, and
  Dolan}]{zhang-etal-2020-pointer}
Yizhe Zhang, Guoyin Wang, Chunyuan Li, Zhe Gan, Chris Brockett, and Bill Dolan.
  2020.
\newblock \href {https://doi.org/10.18653/v1/2020.emnlp-main.698} {{POINTER}:
  Constrained progressive text generation via insertion-based generative
  pre-training}.
\newblock In \emph{Proceedings of the 2020 Conference on Empirical Methods in
  Natural Language Processing (EMNLP)}, pages 8649--8670, Online. Association
  for Computational Linguistics.

\bibitem[{Zhao et~al.(2019)Zhao, Peyrard, Liu, Gao, Meyer, and
  Eger}]{zhao-etal-2019-moverscore}
Wei Zhao, Maxime Peyrard, Fei Liu, Yang Gao, Christian~M. Meyer, and Steffen
  Eger. 2019.
\newblock \href {https://doi.org/10.18653/v1/D19-1053} {{M}over{S}core: Text
  generation evaluating with contextualized embeddings and earth mover
  distance}.
\newblock In \emph{Proceedings of the 2019 Conference on Empirical Methods in
  Natural Language Processing and the 9th International Joint Conference on
  Natural Language Processing (EMNLP-IJCNLP)}, pages 563--578, Hong Kong,
  China. Association for Computational Linguistics.

\bibitem[{Zhu et~al.(2019)Zhu, Hu, and Xing}]{zhu2019text}
Wanrong Zhu, Zhiting Hu, and Eric Xing. 2019.
\newblock \href {https://arxiv.org/abs/1901.00158} {Text infilling}.
\newblock \emph{arXiv preprint arXiv:1901.00158}.

\end{thebibliography}

\newpage
\dpedit{
\section*{Appendix A: DP-based Token Masking Algorithm}

As part of Eq.1 in the main paper, we define the IDF score given token $x_i$ and a corpus $X$ containing $M$ documents as:

\begin{equation*}
    \textrm{IDF}(x_i, X) = -\log \frac{1}{M} \sum_{j=1}^{M} \vmathbb{I} [x_i \in X_j] \ ,
\end{equation*}

\noindent where $\vmathbb{I}[\cdot]$ is the indicator function. We present our DP-based masking algorithm in Algorithm 1:

\begin{algorithm}[h]
\SetAlgoLined
\SetNoFillComment
\KwIn{Human reference $\{x_i\}_{i=1}^N$, masking ratio $\lambda$, and task-specific factor $\alpha$.}
 Compute $v_i$ for each $x_i$ with $\alpha$ (Eq. 1)\;
 Compute $w_i$ depending on LCS for each $x_i$\;
 Init DP-table $T[N+1][W_{\textrm{max}} + 1]$ with all 0\;
 
 \For{$i=1,2,\ldots, N$}{
  \For{$j=1,2,\ldots, W_{\textrm{max}}$}{
  
    \eIf{$j - w_{i-1} < 0$}{
        $T[i][j] = T[i-1][j]$\;
        Record masking choice $\phi(x_i)$\;
    }
    {
        $T[i][j] = \textrm{max}(T[i-1][j],$ \\ $T[i-1][j-w_{i-1}] + v_{i-1})$\;
        Record masking choice $\phi(x_i)$\;
    }
    
  }
 }
 $\{\phi(x_i)_{i=1}^N\} \leftarrow$ backtracking via records\;
 \Return best masking strategy $\{\phi(x_i)_{i=1}^N\}$\;
 \caption{DP-based Token Masking}
\end{algorithm}

\section*{Appendix B: Generate, Judge, and Revise Algorithm}
\label{suB:generate_judge_revise}

The complete procedure for augmenting human references is presented in Algorithm 2. For a given template, we first group the tokens into a block-by-block form with blank blocks (\texttt{[B]}) and text blocks (\texttt{[T]}). Then, we \textit{generate} varying lengths of tokens, iteratively concatenating them with next text block, and \textit{judging} them based on PPL, and finally \textit{revising} current generations accordingly. We use the language modeling ability of LM to check the perplexity of the current sequence, and set a hyper-parameter $\sigma$ to control the maximum extended generation (for a lower PPL).

\begin{algorithm}[!h]
\SetAlgoLined
\SetNoFillComment
\KwIn{Template $\{\phi(x_i)\}_{i=1}^N$, max guess $\sigma$, and LM perplexity checker PPL.}
 Group $\{\phi(x_i)\}_{i=1}^N$ into \texttt{[B]} and \texttt{[T]}\;
 Init final output $s$\;
 \ForEach{\textup{block}}{
    $i \leftarrow 0$\;
    Init priority queue $q$, buffer $s'$\;
    \uIf{\textup{\texttt{[T]}}}{
        Append \texttt{[T]} to $s$\;
  }
  \ElseIf{\textup{\texttt{[B]}}}{
        
        \While{$i < \sigma + |\textup{\texttt{[B]}}|$}{
            
            \eIf{\textup{next is \texttt{[T]}}}{
                $w \leftarrow$ self-planning gen.\;
            }{
                $w \leftarrow$ open-ended gen.\;
            }
            
            $s' \leftarrow s + w$\;
            Record (PPL($s'$ + \texttt{[T]}), $s'$) in $q$\;
            $i \leftarrow i+1$\;
        }
        $s \leftarrow s\ +$ lowest PPL $s'$ pop from $q$\;
  }
 }
 \Return augmented reference $s$\;
 \caption{Generate, Judge and Revise}
\end{algorithm}

Depending on whether there is a subsequent text block, the generation will switch between two modes: self-planning generation (if there is future context) and open-ended generation (otherwise). We use a priority queue to store each step generation and its corresponding PPL for quick revisions afterwards.

}

\end{document}